\documentclass{article}

\usepackage{microtype}
\usepackage{graphicx}
\usepackage{booktabs} 

\usepackage{hyperref}

\usepackage{times}
\usepackage[accepted]{icml2020_arxiv}

\usepackage[utf8]{inputenc} 
\usepackage[T1]{fontenc}    
\usepackage{url}            
\usepackage{amsfonts}       
\usepackage{amsmath}
\usepackage{amssymb}
\usepackage{nicefrac}       
\usepackage{microtype}      
\usepackage{color}
\usepackage{subcaption}
\usepackage{xspace}
\usepackage{stmaryrd}
\usepackage{comment}
\usepackage{afterpage}
\usepackage[normalem]{ulem}
\usepackage{array}

\usepackage{tikz}
\usetikzlibrary{arrows,shapes,automata,backgrounds,through,shadows,fit,positioning}

\tikzstyle{cont}=[circle, draw,thick,minimum size=7.5mm,line width=1pt,>=stealth,positioning]  

\tikzstyle{obs}=[fill=blue!10,thick]  

\tikzstyle{var}=[circle,thick,minimum size=3pt,draw=black,>=stealth,inner sep=2pt, font=\small]

\tikzstyle{latent}  =[]

\tikzstyle{observed}=[fill=gray!50]

\tikzstyle{dgraph}=[->, line width=1.5pt]

\tikzstyle{background}=[rectangle,fill=gray!20,inner sep=0.1cm,rounded corners=5mm]

\tikzstyle{nnet}=[draw, thick, rectangle, minimum height = {height("ReLU")},minimum width={width("BatchNorm+ReLU")}, inner sep=0.2cm, rounded corners=2mm]
    
\tikzstyle{conv}=[fill=orange!20]
\tikzstyle{batch_relu}=[fill=blue!20]

\def\equaldef{\overset{\text{def}}{=}}
\def\NAME{{\sc StackedWAE}}

\def\cX{\mathcal{X}}
\def\cZ{\mathcal{Z}}
\def\cP{\mathcal{P}}
\def\cD{\mathcal{D}}
\def\cN{\mathcal{N}}
\def\cI{\mathcal{I}}

\newcommand{\eqn}[1]{\begin{align}#1\end{align}}
\newcommand{\eqna}[1]{\begin{align}\begin{aligned}#1\end{aligned}\end{align}}

\newcommand{\norm}[1]{\left\lVert#1\right\rVert_{L^2}}

\title{Learning Deep-Latent Hierarchies by Stacking Wasserstein Autoencoders}

\author{
  Benoit Gaujac\\
  University College London\\
    \And
  Ilya Feige\\
  Faculty\\
    \And
  David Barber\\
  University College London\\
}

\begin{document}

\twocolumn[
\maketitle

\vskip 0.3in
]




\begin{abstract}
Probabilistic models with hierarchical-latent-variable structures provide state-of-the-art results amongst non-autoregressive, unsupervised density-based models. However, the most common approach to training such models based on Variational Autoencoders (VAEs) often fails to leverage deep-latent hierarchies; successful approaches require complex inference and optimisation schemes. Optimal Transport is an alternative, non-likelihood-based framework for training generative models with appealing theoretical properties, in principle allowing easier training convergence between distributions. In this work we propose a novel approach to training models with deep-latent hierarchies based on Optimal Transport, without the need for highly bespoke models and inference networks. We show that our method enables the generative model to fully leverage its deep-latent hierarchy, avoiding the well known ``latent variable collapse'' issue of VAEs; therefore, providing qualitatively better sample generations as well as more interpretable latent representation than the original Wasserstein Autoencoder with Maximum Mean Discrepancy divergence.
\end{abstract}

\section{Introduction}
\label{sec:intro}

Probabilistic latent-variable modelling is widely applicable in machine learning as a method for discovering structure directly from large, unlabelled datasets. Variational Autoencoders (VAEs) \citep{kingma2014stochastic,rezende2014stochastic} have proven to be effective for training generative models parametrised by powerful neural networks, by mapping the data into a low-dimensional embedding space. While allowing the training of expressive models, VAEs often fail when using deeper hierarchies with several stochastic latent layers. 

In fact, many of the most successful probabilistic latent-variable models use only one stochastic latent layer. Auto-regressive models \citep{larochelle11a,oord16,PixelCNN,Pixelcnn++}, for example, produce state-of-the-art samples and likelihood scores. However, auto-regressive models suffer from poor scalability to high-dimensional data. Flow-based models \citep{Rezende:2015:VIN:3045118.3045281,NIPS2016_6581,DinhSB16,NIPS2018_8224} are another class of generative models providing competitive sample quality and are able to scale to higher-dimensional data, but they still lag behind auto-regressive models in terms of likelihood scores. 

Deep-latent-variable models are highly expressive models that aim to capture the structure of data in a hierarchical manner, and could thus potentially compete with auto-regressive models for state-of-the-art performance. However, they remain hard to train. Many explanations have been proposed for this, from the use of dissimilarity measure directly in the pixel space \citep{larsen16} resulting in poor sample quality, to the lack of efficient representation in the latent \citep{LearnHierarchFeats}, to simply the poor expressiveness of the models used \citep{ZhaoSE17a,BIVA}. 

Solutions for training deep-latent-variable models range from introducing auxiliary variables to increase the expressiveness of the posterior distribution \citep{pmlr-v48-maaloe16} to replacing the simple spherical Gaussian prior with richer prior distributions which can be learned jointly with the generative model \citep{VampPrior, NIPS2019_8553}. Other approaches focus on building and training complex generative models and inference networks introducing latent skip connections in the generative model and inference network \citep{NIPS2016_6141}, sharing generative model and inference network parameters \citep{LadderVAE}, as well as bidirectional inference networks \citep{BIVA}. \citet{BIVA} managed to train very deep-hierarchical-latent models achieving near state-of-the-art sample generations, both in term of likelihood score and sample quality. However, in order to leverage their latent hierarchy (working in the VAE framework), they needed both complex, tailored inference networks, and deterministic skip connections in the generative model.

Optimal Transport (OT) \citep{villani2008optimal} is a mathematical framework to compute distances between distributions. Recently, it has been used as a training method for generative models \citep{genevay18a,Bousquetetal17}. \citet{WAE} introduced Wasserstein Autoencoders (WAEs), where as with VAEs, an encoding distribution maps the data into a latent space, aiming to learn a low-dimensional representation of the data. WAEs provide a non-likelihood based framework with appealing topological properties \citep{WGAN,Bousquetetal17}, in theory allowing for easier training convergence between distributions. \citet{gaujac2018gaussian} trained a two-latent-layer generative model using a WAE, showing promising results in the capacity of the WAE framework to leverage a latent hierarchy relative to the equivalent VAE.

Following these early successes, we propose to train deep-hierarchical-latent models using the WAE framework, without the need for complex dependency paths in both the generative model and inference network. As in the works of \citet{LadderVAE} and \citet{BIVA}, we believe that a deep-latent hierarchy offers the potential to improve generative models, if they could be trained properly. In order to leverage the deep-latent hierarchy, we derive a novel objective function by stacking WAEs, introducing an inference network at each level, and encoding the information up to the deepest layer in a bottom-up way. For convenience, we refer to our method as \NAME. 

Our contributions are two-fold: first, we introduce \NAME{}, a novel objective function based on OT, designed specifically for generative modelling with latent hierarchies, and show that it is able to fully leverage its hierarchy of latents. Second, we show that \NAME{} performs significantly better when training hierarchical-latent models than the original WAE framework regularising the latent with the Maximum Mean Discrepancy (MMD) divergence \citep{MMD}.

\section{Stacked WAE}
\label{sec:stacked_wae}

\NAME s are probabilistic-latent-variable models with a deep hierarchy of latents. They can be minimalistically simple in their inference and generative models, but are trained using OT in a novel way. We start by defining the probabilistic models considered in this work, then introduce OT, and finally discuss how to train probabilistic models with deep-latent hierarchies using OT, the method that we refer to as \NAME.

Throughout this paper, we use calligraphic letters (e.g. $\cX$) for sets, capital letters (e.g. $X$) for random variables, and lower case letters (e.g. $x$) for their realisation. We denote probability distributions with capital letters (e.g. $P(X)$) and their densities with lower case letters (e.g. $p(x)$).

\subsection{Generative models with deep latent hierarchies}
\label{sec:deep_gen}

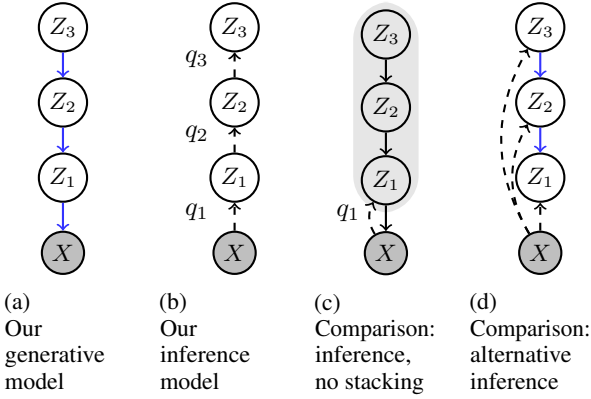
\begin{figure}
\begin{center}
    \begin{subfigure}{0.09\textwidth}
    \centering
        \begin{tikzpicture}[node distance=1.cm]
            \node[var,latent] (z3) {$Z_3$};
            \node[var,latent] (z2) [below of=z3] {$Z_2$};
            \node[var,latent] (z1) [below of=z2] {$Z_1$};
            \node[var,observed] (x) [below of=z1] {$X$};
            \path[->]
                (z3)edge[thick,draw=blue!80](z2)
                (z2)edge[thick,draw=blue!80](z1)
                (z1)edge[thick,draw=blue!80](x);
        \end{tikzpicture}
        \caption{\\Our \\generative model}
        \label{fig_mnist:gen_model}
    \end{subfigure}
\hspace{.02\textwidth}
    \begin{subfigure}{0.09\textwidth}
    \centering
        \begin{tikzpicture}[node distance=1.cm]
            \node[var,latent] (z3) {$Z_3$};
            \node[var,latent,label={[xshift=-.5cm, yshift=0.01cm]$q_3$}] (z2) [below of=z3] {$Z_2$};
            \node[var,latent,label={[xshift=-.5cm, yshift=0.01cm]$q_2$}] (z1) [below of=z2] {$Z_1$};
            \node[var,observed,label={[xshift=-.5cm, yshift=0.01cm]$q_1$}] (x) [below of=z1] {$X$};
            \path[->]
                (z2)edge[thick,dashed](z3)
                (z1)edge[thick,dashed](z2)
                (x)edge[thick,dashed](z1);
        \end{tikzpicture}
        \caption{\\Our \\inference model}
        \label{fig_mnist:inf_model}
    \end{subfigure} 
\hspace{.02\textwidth}
    \begin{subfigure}{0.09\textwidth}
    \centering
        \begin{tikzpicture}[node distance=.97cm]
            \node[var,latent] (z3) {$Z_3$};
            \node[var,latent] (z2) [below of=z3] {$Z_2$};
            \node[var,latent] (z1) [below of=z2] {$Z_1$};
            \node[var,observed,label={[xshift=-.5cm, yshift=0.01cm]$q_1$}] (x) [below of=z1] {$X$};
            \path[->]
                (z3)edge[thick,](z2)
                (z2)edge[thick](z1)
                (z1)edge[thick](x)
                (x)edge[thick,bend left,dashed](z1);
          \begin{pgfonlayer}{background}
                \node [background, fit=(z3) (z1),] {};
            \end{pgfonlayer}                
        \end{tikzpicture}
        \caption{\\Comparison: inference, no stacking}
        \label{fig_mnist:imp_1el}
    \end{subfigure}
\hspace{.02\textwidth}
    \begin{subfigure}{0.09\textwidth}
    \centering
        \begin{tikzpicture}[node distance=1.cm]
            \node[var,latent] (z3) {$Z_3$};
            \node[var,latent] (z2) [below of=z3] {$Z_2$};
            \node[var,latent] (z1) [below of=z2] {$Z_1$};
            \node[var,observed] (x) [below of=z1] {$X$};
            \path[->]
                (x)edge[thick,bend left,dashed](z3)
                (z3)edge[thick,draw=blue!80](z2)
                (x)edge[thick,bend left,dashed](z2)
                (z2)edge[thick,draw=blue!80](z1)
                (x)edge[thick,dashed](z1);
        \end{tikzpicture}
        \caption{\\Comparison: alternative inference}
        \label{fig_mnist:skip_q}
    \end{subfigure}
\caption{(a) Generative model (blue lines represent generative model parameters). (b) Inference model used in \NAME. (c) Standard WAE; 
the generative model has only one latent with prior $p(z_1)=\int p(z_1\vert{}z_2)p(z_2\vert{}z_3)p(z_3)dz_1dz_2dz_3$. (d) Inference model with skips connections and parameter sharing with the generative model, as in \citet{LadderVAE}.}
\label{fig_mnist:graphical_HWAE}
\end{center}
\end{figure}

We will consider deep-generative models with Markovian hierarchies in their latent variables. Namely, where each latent variable depends exclusively on the previous one. Denoting by $P_\Theta$ the parametric model with $N$ latent layers, where $\Theta=\{\theta_1,\dots,\theta_N\}$, we have:
\eqn{
p_\Theta(x,z_{1:N}) = p_{\theta_1}(x|z_1) \Bigg[ \prod_{n=2}^{N} p_{\theta_n}(z_{n-1}|z_{n}) \Bigg]
p_{\theta_N}(z_N)
\label{eq:gen_model}
}
where the data is $X\in\cX$ and the latent variables are $Z_n\in\cZ_n$ and we chose $p_{\theta_N}(z_N)=\cN(z_N;\,0_{\cZ_N},\cI_{\cZ_N})$. The corresponding graphical model for $N=3$ is given Figure~\ref{fig_mnist:gen_model}.

We will be using variational inference through the WAE framework of \citet{WAE}, introducing variational distributions, $q_\Phi(z_1,\ldots,z_N|x)$, to approximate the intractable posterior, analogous to VAEs. It will be shown in Section~\ref{sec:stacking_waes} that without loss of generality, $q_\Phi(z_1,\ldots,z_N|x)$ can have a Markovian latent hierarchy when following the \NAME{} approach. That is, without loss of generality,
\eqn{
q_\Phi(z_{1:N}|x) = 
q_{\phi_1}(z_1|x) \prod_{n=2}^N q_{\phi_n}(z_n|z_{n-1})
\label{eq:inf_model}
}
where each $q_{\phi_n}(z_n|z_{n-1})$ is introduced iteratively by stacking WAEs at each latent layer. The corresponding graphical model for $N=3$ is given Figure~\ref{fig_mnist:inf_model}.

We focus on this simple Markovian latent-variable structure for the generative model 
as a proof point for \NAME. This simple modelling setup is famously difficult to train, as is discussed extensively in the VAE framework (see for example \citet{ImpWeiAE,LadderVAE,LearnHierarchFeats}). 
The difficulty in training such models comes from the Markovian latent structure of the generative model; in particular, the difficulty of learning structure in the deeper latents. This is because, to generate samples $x\sim p_\Theta(x)$, only the joint $p_\Theta(x,z_1)$ is needed as 
$p_\Theta(x)=\int_{\cZ_1}p_{\theta_1}(x|z_1)p_\Theta(z_1)dz_1$.
Learning a smooth structure in each latent layer is not a strict requirement for learning a good generative model, however, it is sought after if the latent is to be used downstream or interpreted. We find empirically (see Section~\ref{sec:mnist}) that a better generative model is also achieved when the latent hierarchy is well learnt all the way down.  

\citet{LadderVAE} sought to overcome the difficulty of learning a deep-latent hierarchy by using deterministic bottom-up inference, followed by top-down inference that shared parameters with the generative model. With additional optimisation tricks (e.g. KL annealing), their deeper-latent distributions would still go unused for sufficiently deep-latent hierarchies  (discussed in \citet{BIVA}). In order to get deeper hierarchies of latents, \citet{BIVA} introduced additional deterministic connections in the generative model as well as bidirectional inference network to facilitate the deep information flow needed to ensure the usage of the deeper-latent layers.

While the approach in \citet{BIVA} is well motivated and achieves excellent results, we choose the OT framework for training deep-latent models due to its topological properties (see \citet{WGAN}). Still, the standard WAE encounters the same difficulties as the VAE in learning deep-latent hierarchies. We thus modify the original WAE objective, effectively stacking WAEs, to improve the learning of both the generative model and the inference distribution throughout the latent hierarchy.

\subsection{Wasserstein Autoencoders}
\label{sec:wae}

The Kantorovich formulation of the OT between the true-but-unknown data distribution $P_D$ and the model distribution $P_\Theta$, for a given cost function $c$, is defined by:
\eqn{
\text{OT}_c(P_D,P_\Theta) = \, \underset{\Gamma\in\cP(P_D,P_\Theta)}{\inf} \int_{\cX\times\cX} c(x,\tilde x) \, d\Gamma(x,\tilde x)
\label{eq:OT}
}
where $\cP(P_D,P_\Theta)$ is the space of all couplings of $P_D$ and $P_\Theta$; namely, the space of joint distributions $\Gamma$ on $\cX\times\cX$ whose densities $\gamma$ have marginals $p_D$ and $p_\Theta$:
\eqn{\label{eq:marginal_constraints}
&\cP(P_D,P_\Theta) \\ \notag
&= \bigg\{ \Gamma \,\Big|
\int_{\cX} \!\! \gamma(x,\tilde x) \,d\tilde x = p_D(x), 
\int_{\cX} \!\! \gamma(x, \tilde x) \,dx = p_\Theta(\tilde x)
\bigg\} 
}

In the WAE framework, 
the space of couplings is constrained to joint distributions $\Gamma$, with density $\gamma$, of the form:
\eqn{
\gamma(x,\tilde x) = \int_{\cZ_{1:N}} p_\Theta(\tilde x|z_{1:N}) \, q_{\phi_1}(z_{1:N}|x) \, p_D(x) \,d z_{1:N}
\label{eq:transportplan_0}
}
where $q_{\phi_1}(z_{1:N}|x)$, for $x\in\cX$, plays the same role as the variational distribution in variational inference.

Marginalising over $\tilde x$ in Eq.~\eqref{eq:transportplan_0} automatically gives $p_D(x)$, thus satisfying the first marginal constraint of Eq.~\eqref{eq:marginal_constraints}. However the second marginal constraint in Eq.~\eqref{eq:marginal_constraints} (that over $x$ giving $p_\Theta$) is not guaranteed. Due to the Markovian structure of the generative model, a sufficient condition for satisfying the second marginal constraint in Eq.~\eqref{eq:marginal_constraints} for all $z_1\in\cZ_1$, is (more detail provided in Appendix~\ref{app:marg_const}):
\eqn{
\int_\cX q_{\phi_1}(z_1 | x) \, p_D(x) \, dx 
= 
p_{\Theta_{2:N}}(z_1)
\label{eq:marginal_const} 
}
where we denote $p_{\Theta_{2:N}}(z_1)$ the prior over $Z_1$: 
\eqn{
p_{\Theta_{2:N}}(z_1) \equaldef \int_{\cZ_{2:N}} \Bigg[ \prod_{n=2}^{N} p_{\theta_n}(z_{n-1}|z_{n}) \Bigg]
p_{\theta_N}(z_N)\, dz_{2:N}
\label{eq:z1_prior}
}

Finally, to get the WAE objective of \citet{WAE}, the constraint in Eq.~\eqref{eq:marginal_const} is relaxed using a Lagrange multiplier (more detail provided in Appendix~\ref{app:wae_obj}):
\eqn{
\widehat{W}_c(P_D,P_\Theta) =& \underset{Q_{\phi_1}(Z_1|X=x)}{\inf}
\bigg[ \int_{\cX\times\cX} \!\! c(x,\tilde x) \, \gamma(x,\tilde x)
\, dx d\tilde{x} 
\notag \\ \label{eq:wae_0} 
&
+
\lambda_1 \, \cD_1\Big(Q_1^\text{agg}(Z_1), \,P_{\Theta_{2:N}}(Z_1)\Big) 
\bigg]
}
where $\cD_1$ is any divergence function, $\lambda_1$ a relaxation parameter, $\gamma$ is defined in Eq.~\eqref{eq:transportplan_0}, and $Q_1^\text{agg}(Z_1)$ is the aggregated posterior:
\eqn{
Q_1^\text{agg}(Z_1) \equaldef \int_{\cX} Q_{\phi_1}(Z_1|x) \,p_D(x) \,dx
\label{eq:Qagg1}
}
Note that because of the Markovian latent structure of the generative model, 
\eqn{
\gamma(x,\tilde x) &= \int_{\cZ_{1:N}} p_\Theta(\tilde x|z_{1:N}) \, q_{\phi_1}(z_{1:N}|x) \, p_D(x) \, d z_{1:N} \notag\\
&=\int_{\cZ_1} p_{\theta_1}(\tilde x|z_1) \, q_{\phi_1}(z_1|x) \, p_D(x) \, d z_1 
\label{eq:only_z1}
}
Eq.~\eqref{eq:Qagg1} and \eqref{eq:only_z1} do not depend on $z_{>1}$, so the infimum in Eq.~\eqref{eq:wae_0} is taken only over $q_{\phi_1}(z_1|x)$ instead of the full $q_{\phi_1}(z_1,\ldots,z_N|x)$.

While Eq.~\eqref{eq:wae_0} is in-principle tractable (e.g. for Gaussian $q_{\phi_1}(z_1|x)$ and sample-based divergence function such as MMD), it only depends on the first latent $Z_1$. Thus it will learn only a good approximation for the joint $p_\Theta(x,z_1)=p_{\theta_1}(x|z_1)p(z_1)$, rather than a full hierarchy with each latent living on a smooth manifold. We show empirically in Section~\ref{sec:imp_prior} that Eq.~\eqref{eq:wae_0} is indeed insufficient for learning to use the full hierarchy of latents.

\subsection{Stacking WAEs for deep latent-variable modelling}
\label{sec:stacking_waes}

In the limit $\lambda_1\to\infty$, Eq.~\eqref{eq:wae_0} does not depend on the choice of divergence $\cD_1$. However, given the set of approximations used, a divergence that takes into account the smoothness of the full stack of latents will likely help the optimisation. We now explain how, by using the Wasserstein distance itself for $\cD_1$, we can derive an objective that naturally pairs up inference and generation at every level in the deep-latent hierarchy. After all, the divergence in Eq.~\eqref{eq:wae_0} is between the intractable aggregate distribution $Q_1^\text{agg}(Z_1)$ from which we can only access samples, and an analytically-known distribution $P_{\Theta_{2:N}}(Z_1)$, which is analgous to where we started with the Wasserstein distance between $P_D$ and the full $P_\Theta$.

Specifically, we choose $\cD_1$ in Eq.~\eqref{eq:wae_0} to be the relaxed Wasserstein distance $\widehat{W}_{c_1}$, which following the same arguments as before, requires the introduction of a new variational distribution $Q_{\phi_2}(Z_2|Z_1)$:
\eqn{
&\cD_1 \Big(Q_1^\text{agg}(Z_1), \, P_{\Theta_{2:N}}(Z_1) \Big) \, = \label{eq:div_1} \\
& \hspace{.6cm} \underset{Q_{\phi_2}(Z_2|Z_1=z_1)}{\inf}
\Bigg[\lambda_2 \, \cD_2\Big( Q_2^\text{agg}(Z_2), \, P_{\Theta_{3:N}}(Z_2) \Big) \notag \\
& \hspace{2cm} + \int_{\cZ_{1}\times\cZ_{1}} \! c_1(z_1,\tilde{z}_1) \, \gamma_1(z_1,\tilde{z}_1) \,dz_1 d\tilde{z}_1
\Bigg] \notag
}
where we re-used the notation of Eq.~\eqref{eq:marginal_const}
for the prior $P_{\Theta_{3:N}}$, and similarly with Eq.~\eqref{eq:Qagg1} for the aggregated posterior, 
\eqn{
Q^\text{agg}_2(Z_2) \equaldef \int_{\cZ_1} Q_{\phi_2}(Z_2|z_1) \, q_1^\text{agg}(z_1) \, dz_1
}
The joint is given by
\eqn{
\gamma_1(z_1,\tilde{z}_1) \equaldef \int_{\cZ_2} \!\! p_{\theta_2}(\tilde{z}_1|z_2) \, q_{\phi_2}(z_2|z_1) \, q_1^\text{agg}(Z_1) \, dz_2
} 
And as before, $Q_{\phi_2}(Z_2|Z_1)$ in the inf does not need to provide a distribution over the $z_{>2}$ without loss of generality.

The divergence $\cD_1$ that arose in Eq.~\eqref{eq:wae_0} between two distributions over $Z_1$ is thus mapped onto the latent at the next level in the latent hierarchy, $Z_2$, via Eq.~\eqref{eq:div_1}. This process can be repeated by using $\widehat{W}_{c_2}$ again for $\cD_2$ in Eq.~\eqref{eq:div_1} to get an expression that maps to $Z_3$, requiring the introduction of another variational distribution $Q_{\phi_3}(Z_3|Z_2)$. Repeating this process until we get to the final layer of the hierarchical-latent-variable model gives the \NAME{} objective:
\eqn{
&W_{\text{\NAME}}(P_D,P_\Theta) \,= \label{eq:wae_unrolled_final} \\
& \underset{Q_\Phi(Z_1,\ldots,Z_N|X=x)}{\inf} \Bigg[ \bigg[\prod_{i=1}^N\lambda_i\bigg] \, \cD_{N} \Big( Q_N^\text{agg}(Z_N), \, P(Z_N) \Big) \notag \\
&+ \sum_{n=0}^{N-1} \, \bigg[\prod_{i=1}^n\lambda_i\bigg]
\int_{\cZ_{n}\times\cZ_{n}} c_n(z_n,\tilde{z}_n) \, \gamma_n(z_n,\tilde{z}_n) \, dz_n \, d\tilde{z}_n
\Bigg]  \notag 
}
where we denote $(\cZ_0, Z_0, z_0) = (\cX, X, x)$ and we define the empty product $\prod_{i=1}^0\lambda_i \equaldef 1$. Similarly to Eq.~\eqref{eq:div_1}, we defined the joints $\gamma_n$ as
\eqn{
& \gamma_n(z_n,\tilde{z}_n) \, \equaldef  \\ \notag
& \hspace{.5cm} \int_{\cZ_{n+1}} p_{\theta_{n+1}}(\tilde{z}_n|z_{n+1}) \, q_{\phi_{n+1}}(z_{n+1}|z_n) \, q_n^\text{agg}(z_n) \, dz_{n+1} 
}
Each $p_{\theta_n}$ is the $n^\text{th}$ layer of the generative model given in Eq.~\eqref{eq:gen_model}. The $q_{\phi_n}$'s are the inference models introduced each time a WAE is "stacked", which combine to make the overall \NAME{} Markovian inference model given in Eq.~\eqref{eq:inf_model}, and the aggregated posterior distributions are defined as
\eqn{\label{eq:Qz}
Q_n^\text{agg}(Z_n) \, \equaldef \int_{\cZ_{n-1}} Q_{\phi_n}( Z_n|z_{n-1}) \, q_{n-1}^\text{agg}(z_{n-1}) \, dz_{n-1} 
}
with $Q_0^\text{agg} \, \equaldef \, P_D$.

Note that the \NAME{} objective function in Eq.~\eqref{eq:wae_unrolled_final} still requires the specification of a divergence function at the highest latent layer $\cD_N$, which we simply take to be the MMD as originally proposed by \citet{WAE}. Other choices can be made, as in \citet{SAE}, who choose a Wasserstein distance computed using the Sinkhorn algorithm \citep{NIPS2013_4927}. While \citet{SAE} provide a theoretical justification for the minimisation of a Wasserstein distance in the prior space, we found that it did not result in significant improvement and comes at an extra efficiency cost. Similarly, one could choose different cost functions $c_n$ at each layer; for simplicity we take all cost functions to be the squared Euclidean distance in their respective spaces.

\section{Experiments}
\label{sec:exp}

We now show how the \NAME{} approach of Section~\ref{sec:stacked_wae} can be used to train deep-latent hierarchies without customizing the generative or inference models (e.g. no skip connections, no parameter sharing). We also show through explicit comparison that the \NAME{} approach performs significantly better than the standard WAE when training deep-hierarchical-latent models.

\subsection{MNIST}
\label{sec:mnist}

\subsubsection*{Experimental setup}
\label{sec:exp_setup}

We trained a deep-hierarchical-latent variable model with $N=5$ latent layers on raw (non-binarised) MNIST \citep{MNIST}. The latent layers have dimensions: $d_{\cZ_1}=32$, $d_{\cZ_2}=16$, $d_{\cZ_3}=8$, $d_{\cZ_4}=4$ and $d_{\cZ_5}=2$. We chose Gaussian distributions for both the generative and inference models, except for the bottom layer of the generative model, which we choose to be deterministic, as in \citet{WAE}. The mean and covariance matrices used are parametrised by fully connected neural networks similar to that of \citet{LadderVAE}. Full details of the experimental setup is given in Appendix~\ref{app:mnist_setup}.

\subsubsection*{Learning a deep-latent hierarchy}
\label{sec:deep_latent_results}

\begin{figure}
\begin{center}
\begin{subfigure}{.22\textwidth}
    \centering\includegraphics[width=\textwidth]{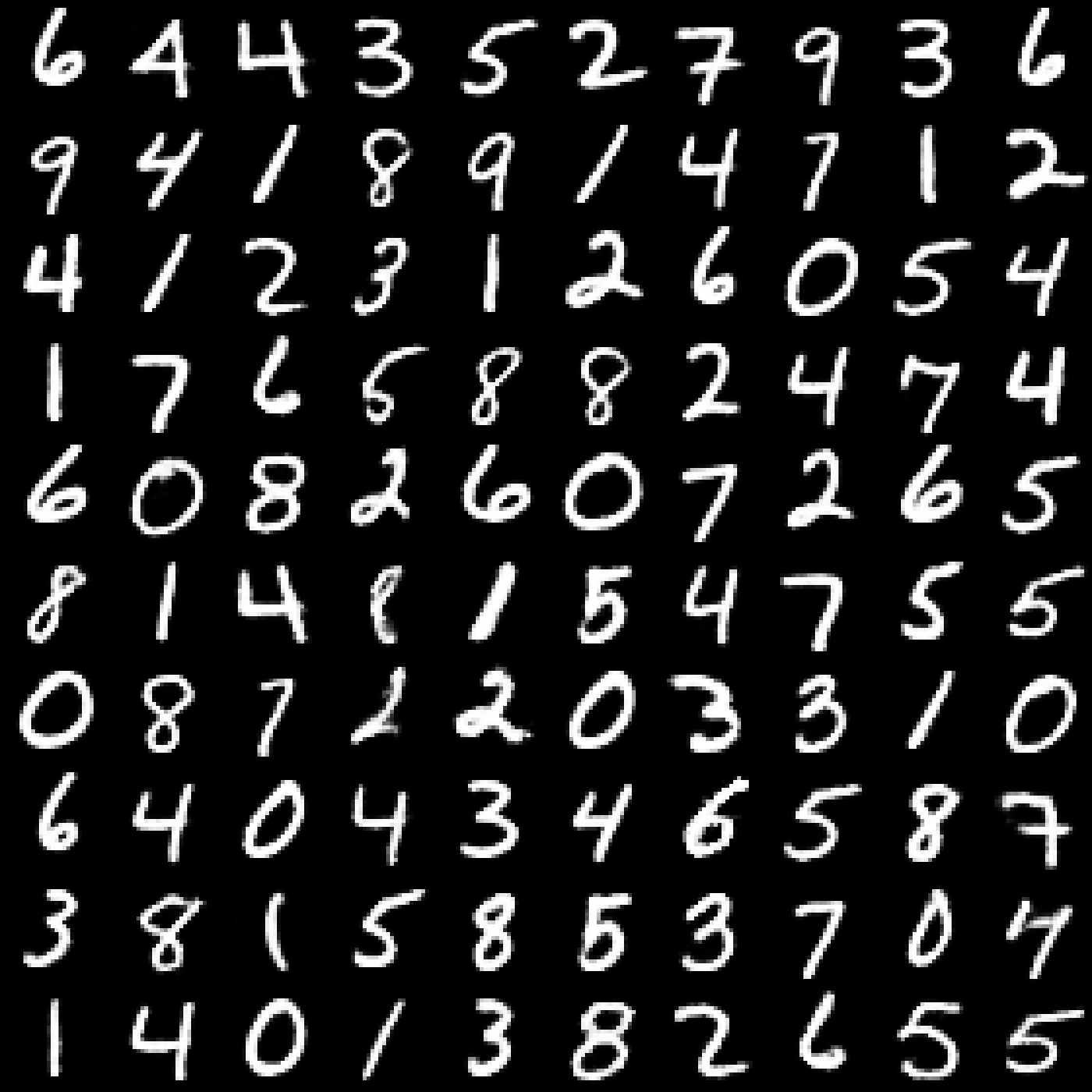}
    \vskip -.3em
    \caption{}
    \vskip -.4em
    \label{fig_mnist:prior_samples}
\end{subfigure}
\hspace{.01\textwidth}
\begin{subfigure}{.22\textwidth}
    \centering\includegraphics[width=\textwidth]{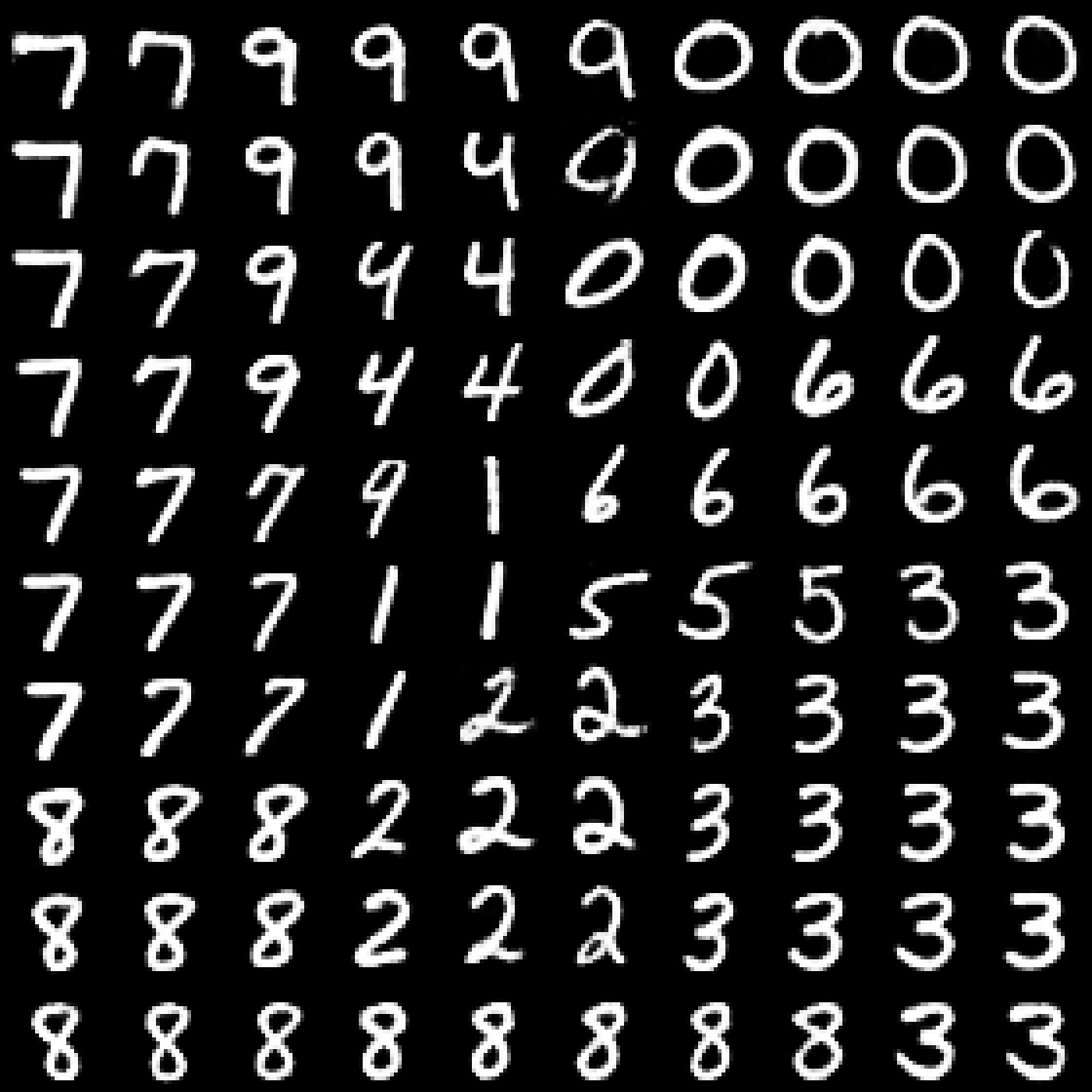}
    \vskip -.3em
    \caption{}
    \vskip -.4em
    \label{fig_mnist:latent_interpolations}
\end{subfigure}
\caption{5-layer \NAME.
(a) Model samples. (b) $\cZ_5$ latent-space interpolations in a 2-dimensional grid varying between $\pm 2$ standard deviations from the origin.}
\label{fig_mnist:mnist_fig1}
\end{center}
\end{figure}

Our results are shown in Figure~\ref{fig_mnist:mnist_fig1}, with samples from the generative model in Figure~\ref{fig_mnist:prior_samples}, and latent space interpolations in Figure~\ref{fig_mnist:latent_interpolations} where digits are reconstructed from points in $\cZ_5$ taken evenly in a grid varying between $\pm 2$ standard deviations from the origin. \NAME{} generates compelling samples and learns a smooth manifold (see Figure~\ref{fig_mnist:app_mnist}, top-left, and Figure~\ref{fig_mnist:point_inter} of Appendix~\ref{app:mnist} for additional samples and latent interpolations respectively). Note that the choice $d_{\cZ_5}=2$ allows for easy visualisation of the learned manifold, rather than being the optimal dimension for capturing all of the variance in MNIST.

\begin{figure}
\begin{center}
\begin{subfigure}{.49\textwidth}
    \centering\includegraphics[width=\textwidth]{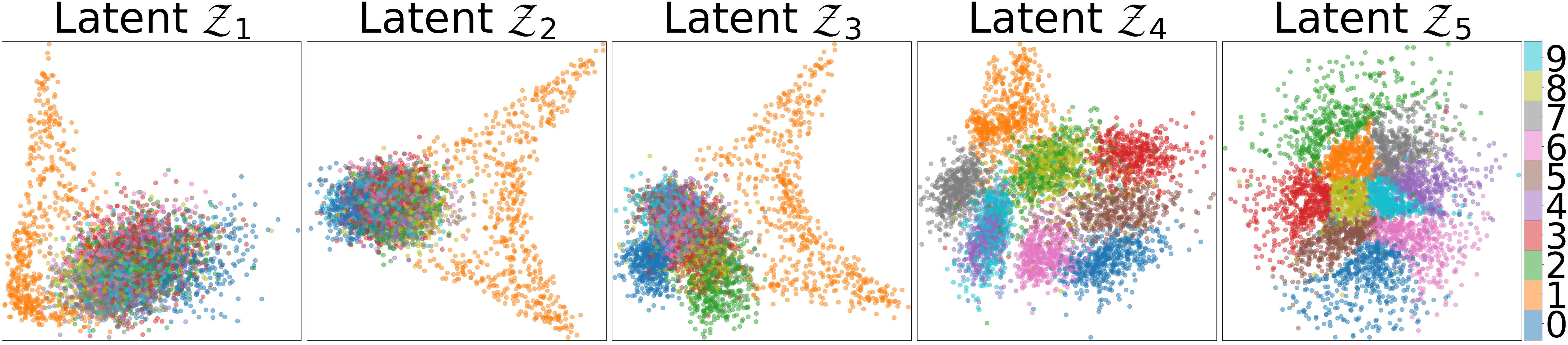}
\vskip -.3em
\caption{}
\label{fig_mnist:pca}
\end{subfigure}\\
\vspace{2.7mm}
\begin{subfigure}{.49\textwidth}
    \centering\includegraphics[width=\textwidth]{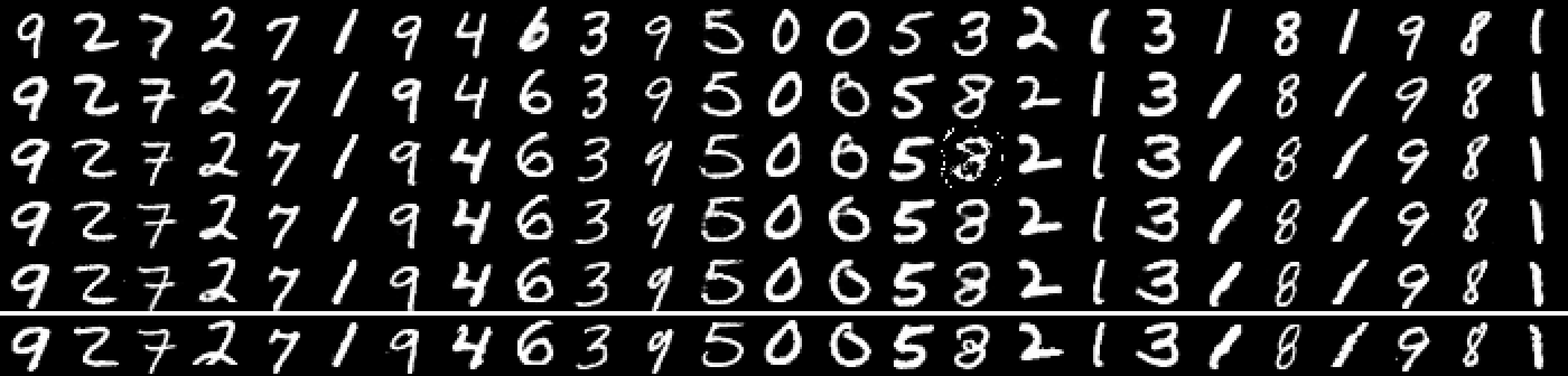}
\vskip -.3em
\caption{}
\vskip -.4em
\label{fig_mnist:full_recon}
\end{subfigure}
\caption{5-layer \NAME. (a) Visualisations of latent spaces $\mathcal{Z}_i$. Each colour corresponds to a digit label. $d_{\cZ_5}=2$ can be directly plotted; for higher dimensions we use PCA. (b) Reconstructions for different encoding layers. The bottom row is data; the $i^\text{th}$ row from the bottom is generated using the latent codes $z_i$ which are from the $i^\text{th}$ encoding layer.}
\end{center}
\end{figure}

Figure~\ref{fig_mnist:latent_interpolations} shows that \NAME{} manages to use all of its latent layers, capturing most of the covariance structure of the data in the deepest-latent layer, which is something that VAE methods struggle to accomplish \citep{LadderVAE,LearnHierarchFeats}. Figure~\ref{fig_mnist:pca} shows the encoded input images through the latent layers, with corresponding digit labels coloured. We see through each layer that \NAME{} leverages the full hierarchy in its latents, with structured manifolds learnt at each stochastic layer. Note that for the shallower layers with higher dimensions (\textit{e.g.} $d_{\cZ_1}=32$ or $d_{\cZ_2}=16$), the PCA algorithm results in a poor visualisation. For such high dimensional spaces, other visualisation techniques can be used such as  UMAP \citep{2018arXivUMAP} as done in Figure~\ref{fig_mnist:umap} of Appendix~\ref{app:mnist_add_res}.

\begin{figure}
\begin{center}
\begin{subfigure}{.22\textwidth}
    \centering\includegraphics[width=\textwidth]{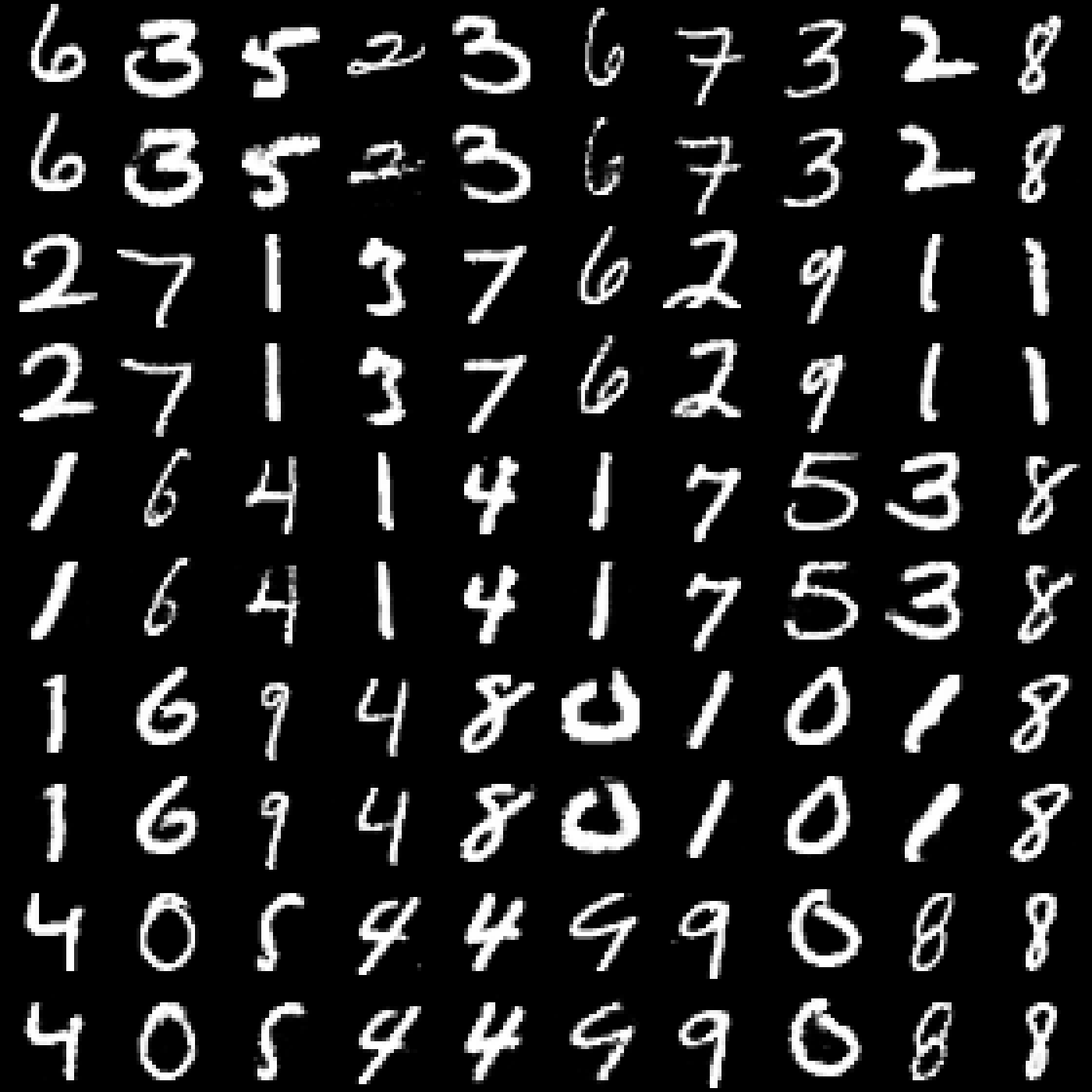}
    \caption{}
    \vskip -.4em
    \label{fig_mnist:imp_recons}
\end{subfigure}
\hspace{.01\textwidth}
\begin{subfigure}{.22\textwidth}
    \centering\includegraphics[width=\textwidth]{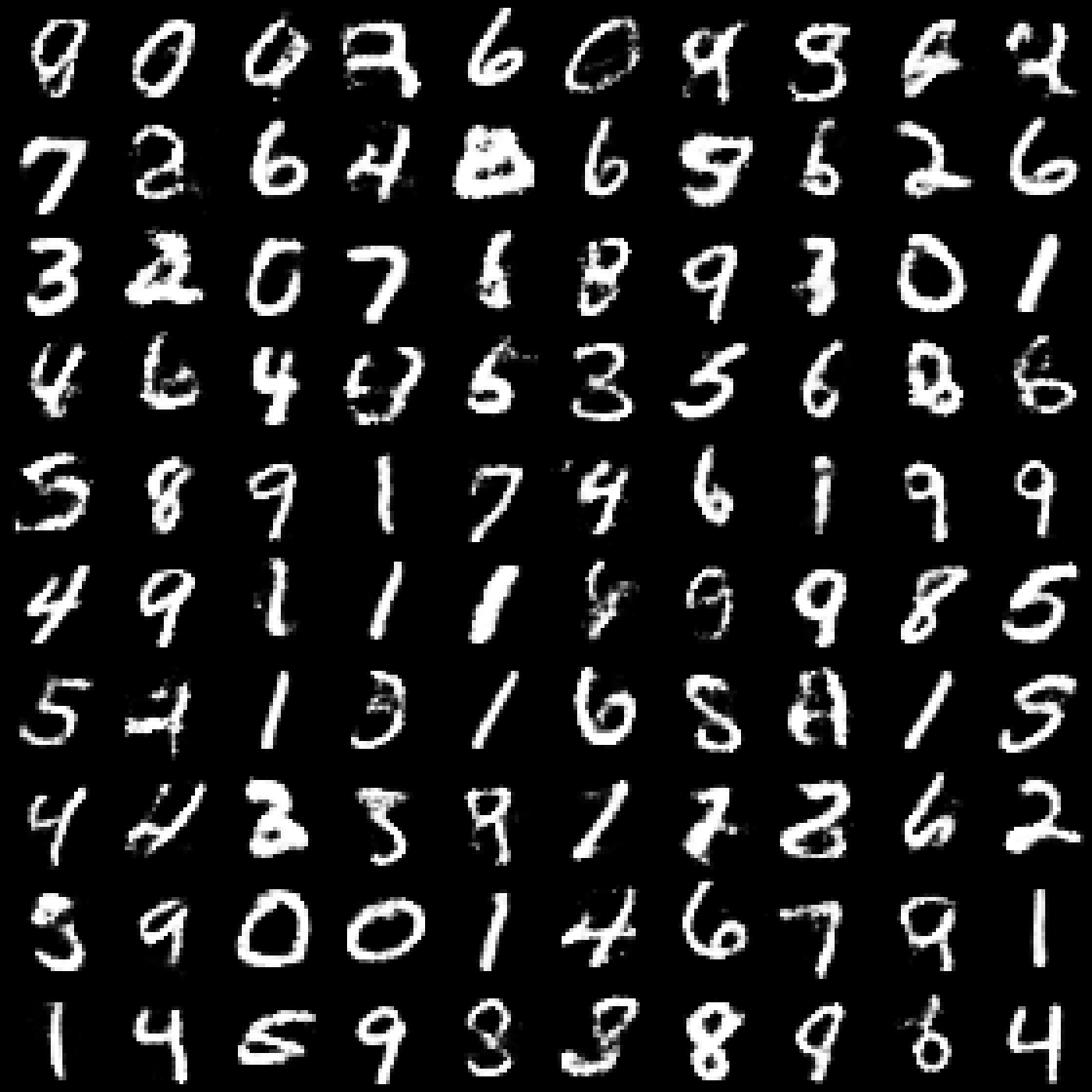}
    \caption{}
    \vskip -.4em
    \label{fig_mnist:imp_latent_inter}
\end{subfigure}
\caption{1-layer implicit-prior WAE. (a) Reconstructions (within pairs of rows, data is above with the corresponding reconstructions below).
(b) $\cZ_5$ latent-space interpolations.}
\label{fig_mnist:mnist_impl}
\end{center}
\end{figure}

An advantage of deep-latent hierarchies is their capacity to capture information at different levels, augmenting a single-layer latent space. In Figure~\ref{fig_mnist:full_recon}, input images, shown in the bottom-row, are encoded and reconstructed for each latent layer. More specifically, the inputs are encoded up to the latent layer $i$ and reconstructed from the encoded $z_i$ using the generative model $p(x\vert{}z_1)p(z_1\vert{}z_2)\ldots p(z_{i-1}\vert{}z_i)$. Row $i$ in Figure~\ref{fig_mnist:full_recon} shows the reconstruction obtained from encoding up to layer $i$. We can see that each additional encoding layer moves slightly farther away from copying the input image as it moves towards fitting the encoding into the 2-dimensional, unit-normal prior distribution. The dimensionality of the deeper-latent layers is a modelling choice which determines how much information is preserved; this can be seen through the loss of information from deeper reconstructions in Figure~\ref{fig_mnist:full_recon} (for a better visualisation of the \textit{real} model reconstructions, corresponding to the top-row in Figure~\ref{fig_mnist:full_recon}, see Figure~\ref{fig_mnist:app_mnist}, bottom-left, of Appendix~\ref{app:mnist}). Indeed, in each layer, the encoder is asked to map the incoming encodings into a lower-dimensional latent space, filtering the amount of information to keep and pass up to the deeper layer. Thus, if there is a mismatch in the dimensions between the true underlying generative process of the data and the chosen model, the encoder will have to project the encodings into lower-dimensional space, losing information along the way.

\subsubsection*{Ablation study: StackedWAE versus WAE}
\label{sec:imp_prior}

In this section, we compare \NAME{} with the original WAE framework for training generative models with deep-hierarchical latents. In particular, we train a WAE using the objective defined in Eq.~\eqref{eq:wae_0} and an inference distribution as in Eq.~\eqref{eq:inf_model}; the corresponding graphical model is shown in Figure~\ref{fig_mnist:imp_1el}. We use the same the experimental setup as outlined earlier in this section, and the same parametrised networks. This experiment can be related to the work of \citet{VampPrior, NIPS2019_8553} in the VAE framework case, as we can rephrase it as training a plain-vanilla 1-layer WAE whose prior over the latent variable is defined and parametrised as in Eq.~\eqref{eq:z1_prior}, and learned alongside the inference network and the generative model. However, we do not use any specific optimisation scheme nor do we constraint the structure of the prior beside the modelling choices made in Section~\ref{sec:deep_gen}, as opposed to what is done in \citet{VampPrior, NIPS2019_8553}.

The results are shown in Figure~\ref{fig_mnist:mnist_impl}, with  reconstructions in Figure~\ref{fig_mnist:imp_recons}, and deepest-latent interpolations in Figure~\ref{fig_mnist:imp_latent_inter}. The latter two should be compared directly with bottom and top-row of Figure~\ref{fig_mnist:full_recon} and Figure~\ref{fig_mnist:latent_interpolations}, respectively, for the \NAME{}  (see also Figure~\ref{fig_mnist:app_mnist} of Appendix~\ref{app:mnist} for samples and reconstructions comparisons with \NAME{}). The samples generated with the standard WAE when training deep-hierarchical-latent variable models (Figure~\ref{fig_mnist:app_mnist}, top-right, of Appendix~\ref{app:mnist}) are poor in comparison with those of the \NAME{} (Figure~\ref{fig_mnist:app_mnist}, top-left, of Appendix~\ref{app:mnist}).

The lack of smooth interpolations in Figure~\ref{fig_mnist:imp_latent_inter} shows that almost no structure has been captured in the deepest-latent layer. 
This is likely due to the fact that the standard WAE, with objective given in Eq.~\eqref{eq:wae_0}, is independent of the deeper-latent inference distributions, thus weakening the smoothness constraint in the deeper layers.
The relatively accurate reconstructions in Figure~\ref{fig_mnist:imp_recons} indicate that the model only needs the first latent layer to capture most of the structure of the data.
This behaviour is similar to that of the Markov HVAE as described in \citet{LearnHierarchFeats}. They show that, for Markov HVAEs to learn interpretable latents, one needs additional constraints beyond simply maximising the likelihood, or in our case, the WAE objective of \citet{WAE} with MMD divergence.

\subsection{Real world Datasets}
\label{sec:svhn_celeba}

We now turn to more realistic datasets to show that \NAME{} is still able to leverage a deep-latent hierarchy. In particular, we trained a 6-layer and a 10-layer \NAME{} on Street View House Numbers (SVHN) \citep{SVHN} and CelebA \citep{liu2015faceattributes}, respectively. Our goal in this section, is to train very deep-latent variable models such as the ones in \citet{BIVA}, and show how \NAME{} still manages to use the deepest-latent layers of the generative model. It is worth stating that we are not aiming to achieve state-of-the-art performance on these datasets, but rather to show that it is easy to learn a deep-hierarchical latent representation of the data.

Our implementation choices resulted in relatively high-dimensional latent spaces (see below and Appendix~\ref{app:svhn_celeba} for more details).
\citet{WAE_followup} observed that when training WAEs with high-dimensional latent spaces, the variance of the inference network tends to collapse to $0$. The authors argue that this might come either from an optimisation issue or from the failing of the divergence used to regularise the latents. Either way, the collapse to deterministic encoders results in poor sample quality as the deterministic encoder is being asked to map the input manifold into a higher-dimensional space than its intrinsic dimension. They proposed to explicitly include a regulariser that maintains a non-zero variance in  Eq.~\eqref{eq:wae_unrolled_final}. As in \citet{WAE_followup}, we included a log-variance penalty term given by Eq.~\eqref{eq:pen_encsigma}:
\eqn{
\mathcal{L}_\text{pen}&=\sum_{i=1}^{N} \lambda^\Sigma_i\sum_{m=1}^{d_{\cZ_i}} |\log \Sigma^q_i[m]|
\label{eq:pen_encsigma}
}
Where $\Sigma^q_i$ is the covariance matrix of the $n^\text{th}$ inference network.
We find that an exponentially decreasing penalty term $\lambda^\Sigma_i$ (see following sections) works well for the dataset at hand with our experimental setup. This choice is motivated by the fact that the bigger the latent dimension (shallower latent layers in the hierarchy), the more likely it is that the latent dimension is larger than the data's intrinsic dimension.

\subsubsection*{Street View House Numbers}
\label{sec:svhn}

We trained a 6-layer \NAME{} with Gaussian inference networks and generative models at each latent layer, with mean and covariance functions parametrised by 3-layer ResNet-style neural networks \citep{resnet}. The details are given in Appendix~\ref{app:svhn}. These architectures dictate the dimensionality of the latent spaces, with the latent dimension given by the size of the output feature map at that layer times the number of these feature maps. In this experiment, networks $i=1,3,5$ have stride 2, effectively performing  downsampling and upsampling operations in the bottom-up path and the top-down path respectively, resulting in latent dimensions ($\text{width}\times\text{height}\times\text{feature map}$) of: 
$16\!\!\times\!\! 16\!\!\times\!\!2$, $16\!\!\times\!\!16\!\!\times\!\!1$, $8\!\!\times\!\!8\!\!\times\!\!2$, $8\!\!\times\!\!8\!\!\times\!\!1$, $4\!\!\times\!\!4\!\!\times\!\!2$ and $4\!\!\times\!\!4\!\!\times\!\!1$.\\
For the regularisation hyperparameters, we use $\prod_{i=1}^n\lambda_i = \lambda_\text{rec}^{(n-1)+1}$ for $n=1,\ldots,5$ (each reconstruction term in the objective), and $\prod_{i=1}^6\lambda_i = \lambda_\text{match}$ (the final divergence term). The choice for the reconstruction weights is motivated by the fact that the effective regularisation hyperparameters scale exponentially. Thus, to avoid the collapse (or blowing up for $\lambda_\text{rec}>1$) of the corresponding reconstruction terms, we choose the weights to scale like $\mathcal{O} \big(\lambda_\text{rec}^{n}\big)$. We found that $\big(\lambda_\text{rec},\lambda_\text{match}\big)=\big(0.5,10\big)$ worked well with our experimental setup.
As mentioned above, in order to avoid the collapse to deterministic encoder networks, we penalised the log-variance of the inference  networks (as done in \citet{WAE_followup}). We found that $\lambda^\Sigma_i=\lambda^\Sigma\!\cdot \text{e}^{-(i-1)}$, for $i=1,\ldots,6$, with $\lambda^\Sigma=2.5$ worked well in our setting. More details of the experimental setup can be found in Appendix~\ref{app:svhn}.

\begin{figure}
\begin{center}
\begin{subfigure}{.48\textwidth}
    \centering\includegraphics[width=\textwidth]{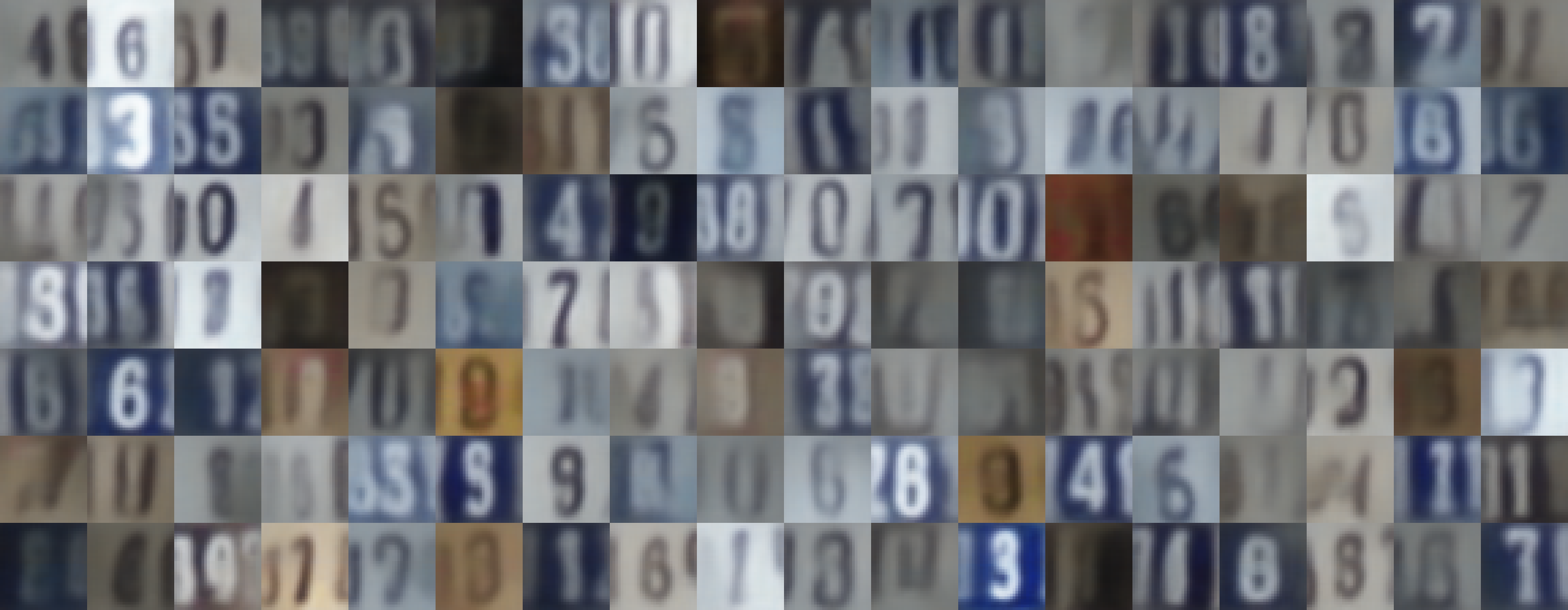}
    \vskip -.1em
    \caption{}
    \label{fig_svhn:prior_samples}
\end{subfigure}\\
\vspace{3.5mm}
\begin{subfigure}{.48\textwidth}
    \centering\includegraphics[width=\textwidth]{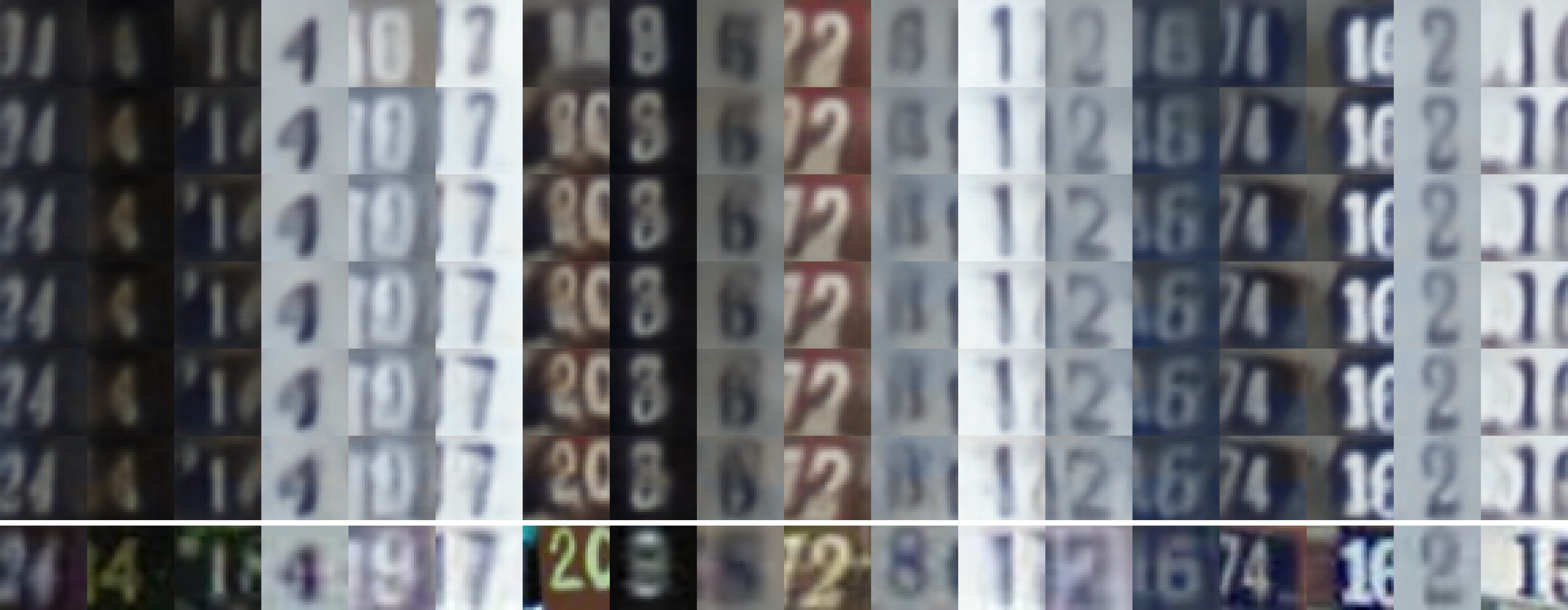}
    \vskip -.1em
    \caption{}
    \label{fig_svhn:full_recon}
\end{subfigure}\\
\vspace{3.5mm}
\begin{subfigure}{.48\textwidth}
    \centering\includegraphics[width=\textwidth]{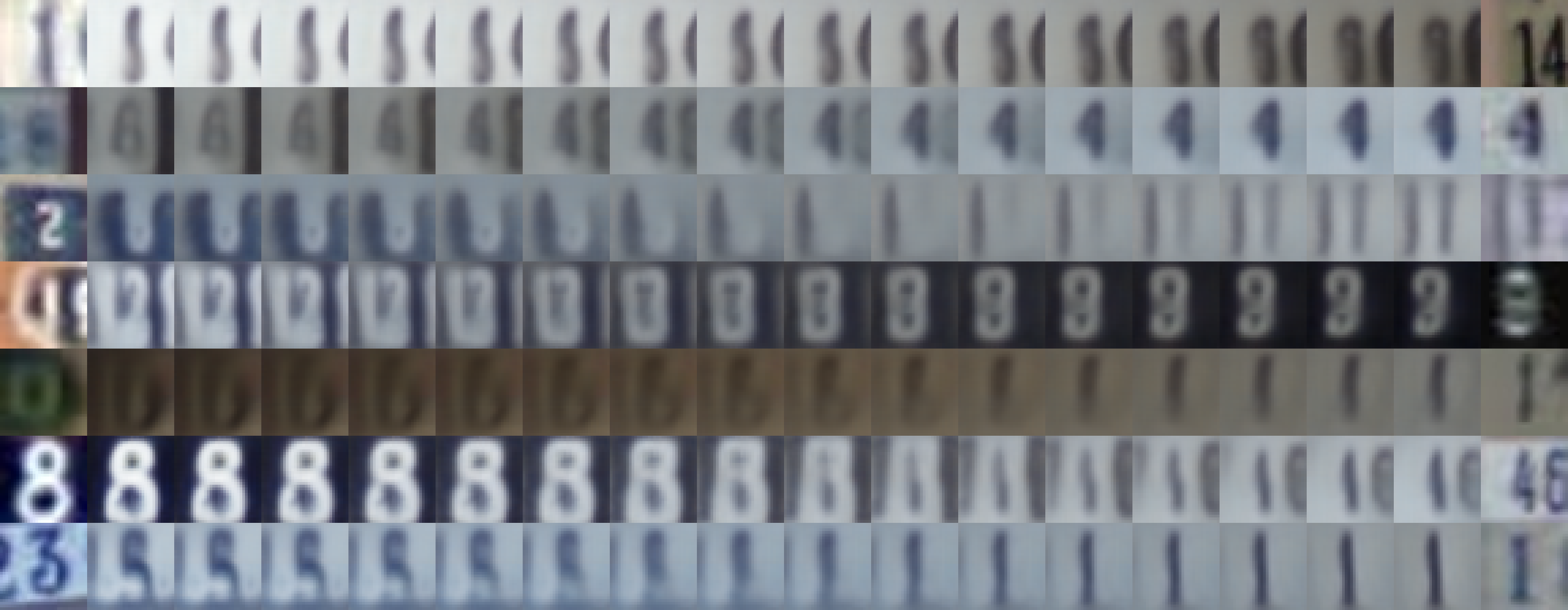}
    \vskip -.1em
    \caption{}
    \vskip -.1em
    \label{fig_svhn:point_inter}
\end{subfigure}
\caption{6-layer \NAME. (a) Model samples. (b) Reconstructions for different encoding layers, as in Figure~\ref{fig_mnist:full_recon}. (c) Points interpolations; the first and last columns are actual data points,
with corresponding reconstructions shown in the second, respectively second-to-last, columns.
}
\label{fig_svhn:svhn_fig1}
\end{center}
\end{figure}

Our results on SVHN are given in Figure~\ref{fig_svhn:svhn_fig1}. Samples from the generative model are shown in Figure~\ref{fig_svhn:prior_samples}.
Figure~\ref{fig_svhn:full_recon} shows the reconstructions of the data points (along the bottom row) at each latent layer in the hierarchy. Similarly to the results for MNIST, we can see that the deepest latent layer may not be large enough to enable high-fidelity reconstructions. Our intention is to show that the hierarchy of latents can be learnt, which is clearly the case, rather than to model SVHN perfectly, so we do not attempt to tune to the optimal latent dimensionality. Figure~\ref{fig_svhn:point_inter} represents point interpolation, with the actual anchor data points shown in the first and last column, their reconstructions in the second, respectively second-to-last, columns, and the latent interpolation in-between.

\subsubsection*{CelebA}

We trained a 10-layers \NAME{}. Similarly to SVHN, the inference and generative networks are fully-convolutional ResNet. The inference networks and generative models, $i=1,4,7,10$ have stride 2. These downsampling, respectively upsampling, operations resulted in 4 groups of latent spaces sharing features maps of the same spatial dimensions: latent layers $i=1,2,3$ with output features maps of size $32\!\!\times\!\!32$, layers $i=4,5,6$ with features maps of size $16\!\!\times\!\!16$, layers $i=7,8,9$ with size $8\!\!\times\!\!8$ and layer $i=10$ with size $4\!\!\times\!\!4$. Within each group, the number of feature maps is decreased by two as we go up the hierarchy, with the biggest layer having $8$ features maps and the smallest $4$. As with SVHN, we used an exponentially decreasing regularisation parameters. More precisely, we chose $\prod_{i=1}^p\lambda_i = \lambda_\text{rec}^{(p-1)/3+1}$ for $p=1,\ldots,9$ and $\prod_{i=1}^{10}\lambda_i = \lambda_\text{match}$. For this dataset, choosing $\big(\lambda_\text{rec},\lambda_\text{match}\big)=\big(10^{-1},10^{0}\big)$ worked well. We also added a $L_1$ penalty on the log-covariance matrices of the encoders to avoid any variance collapse. We used the same penalisation scheme than in the previous experiment, that is $\lambda^\Sigma_i=\lambda^\Sigma\!\cdot \text{e}^{-(i-1)}$, for $i=1,\ldots,10$, with $\lambda^\Sigma=2.5$. See Appendix~\ref{app:celeba} for more details about the experimental setup.

\begin{figure}
\begin{center}
\begin{subfigure}{.49\textwidth}
    \centering\includegraphics[width=\textwidth]{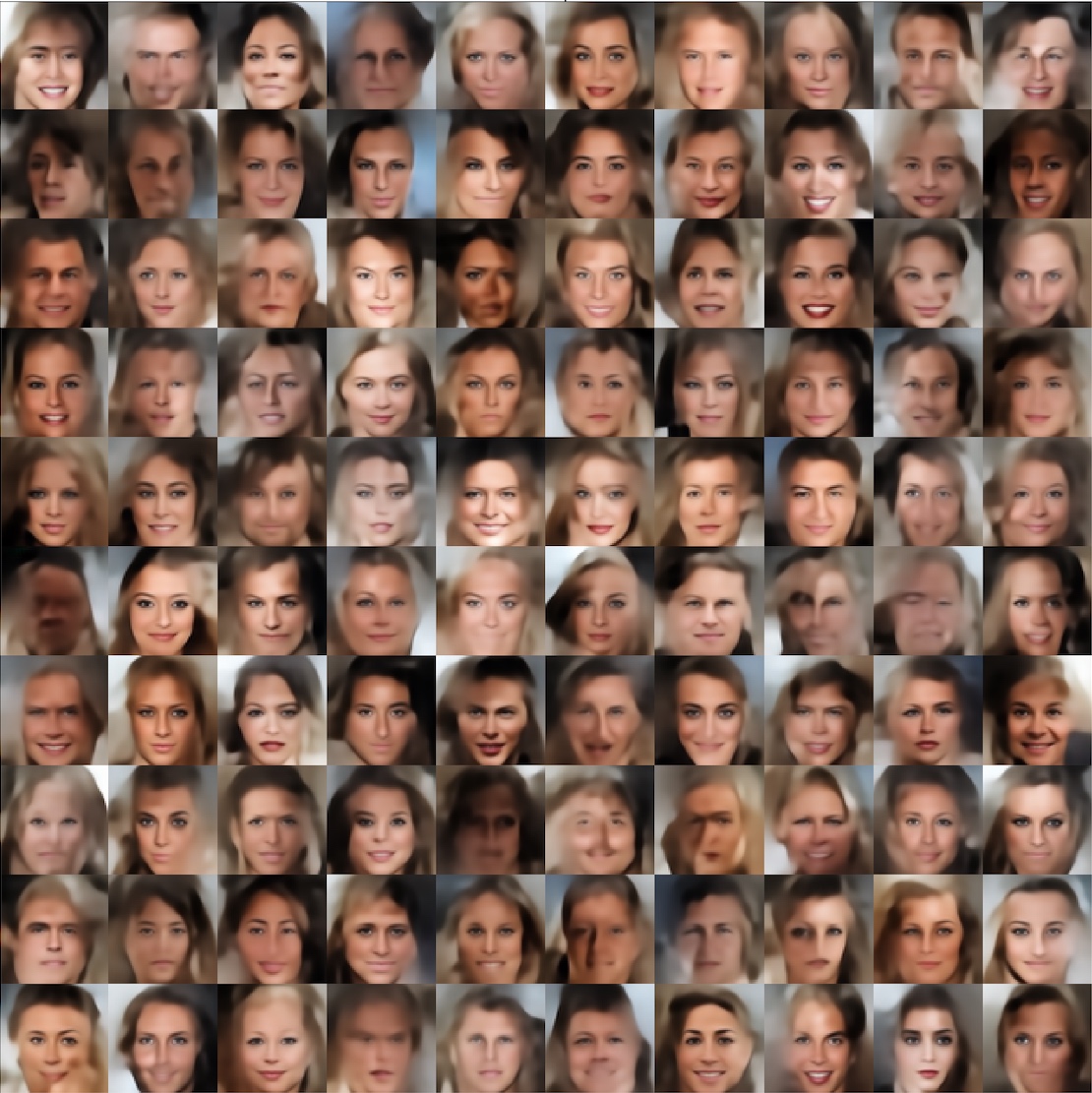}
    \vskip -.2em
    \caption{}
    \label{fig_celeba:prior_samples}
    \vspace{2.8mm}
\end{subfigure}\\
\begin{subfigure}{.49\textwidth}
    \centering\includegraphics[width=\textwidth]{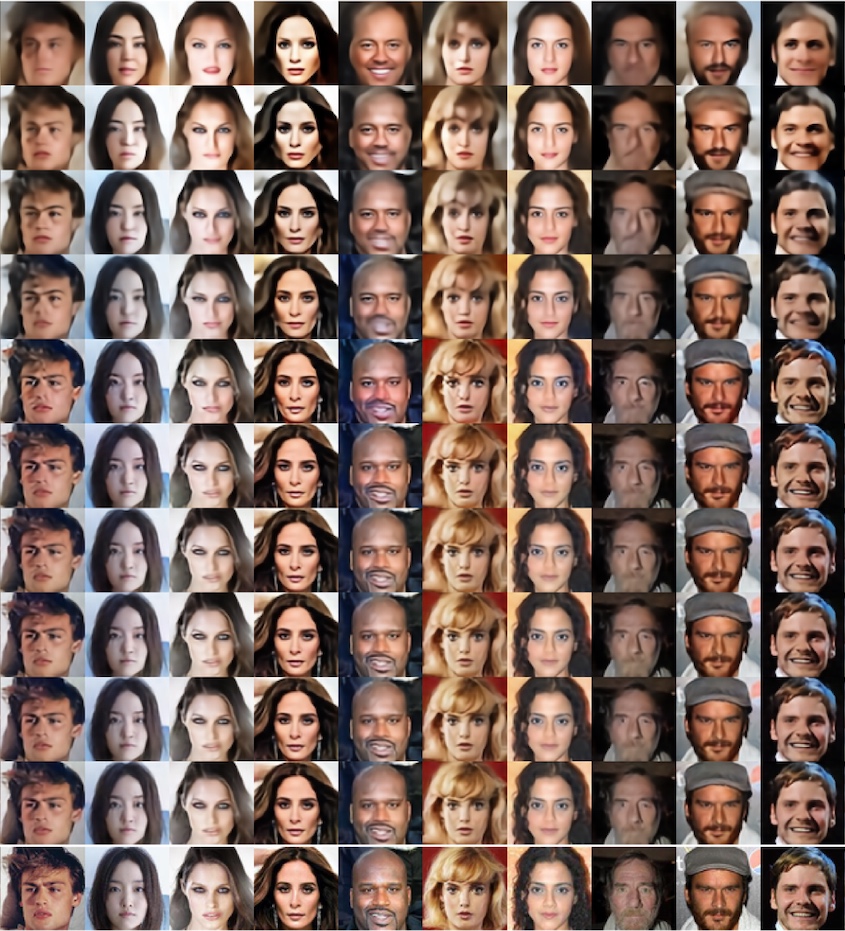}
    \vskip -.2em
    \caption{}
    \vskip -.3em
    \label{fig_celeba:full_recon}
\end{subfigure}
\caption{10-layer \NAME. (a) Model samples.
(b) Reconstructions for different encoding layers, as in Figure~\ref{fig_mnist:full_recon}.}
\label{fig_celeba:celeba_fig1}
\end{center}
\end{figure}

Results are shown Figure~\ref{fig_celeba:celeba_fig1}, with model samples given in Figure~\ref{fig_celeba:prior_samples} and layer-wise reconstructions in Figure~\ref{fig_celeba:full_recon}. As with the 6-layer \NAME{} trained on SVHN, Figure~\ref{fig_celeba:full_recon} shows that \NAME{} manages to use all its latent layers, up to the deepest ones. While the full reconstructions, shown in the top-row of Figure~\ref{fig_celeba:full_recon}, are relatively close to the original data points, shown in the bottom-row, we can notice, as in the MNIST and SVHN experiments, a loss of information as we go deeper in the hierarchy. In other words, the deeper the encoding, the smoother or blurrier the reconstructions. Here again, one possible explanation is the excessive filtering of information by the encoder when going up in the hierarchy due to the miss-match between the intrinsic dimension and the latent dimension.

\section{Conclusion}

In this work we introduced a novel objective function for training generative models with deep hierarchies of latent variables using Optimal Transport. Our approach recursively applies the Wasserstein distance as the regularisation divergence, allowing the stacking of WAEs for arbitrarily deep-latent hierarchies. We showed that this approach enables the learning of smooth latent distributions even in deep latent hierarchies, which otherwise requires extensive model design and tweaking of the optimisation procedure to train. We also showed that our approach is significantly more effective at learning smooth hierarchical latents than the standard WAE.

\bibliography{icml2020_arxiv}
\bibliographystyle{icml2020_arxiv}
\null\newpage


\null\newpage
\appendix
\section{StackedWAE derivation details}
\label{app:stackedwae_obj}

\subsection{Marginal constraint}
\label{app:marg_const}

The space of couplings $\Gamma\in\cP(P_D,P_\Theta)$ that defines the OT distance in Eq.~\eqref{eq:OT} is constrained according to Eq.~\eqref{eq:marginal_constraints}. WAE assumes a joint density of the form given in Eq.~\eqref{eq:transportplan_0}, which automatically satisfies the $p_D$ marginal constraint, but requires the further sufficient condition given in Eq.~\eqref{eq:marginal_const} in order to satisfy the $p_\Theta$ marginal constraint. To see that Eq.~\eqref{eq:marginal_const} is indeed a sufficient condition for the $p_\Theta$ marginal constraint, note that from the Markovian assumption of the generative model (see Eq.~\eqref{eq:gen_model}), we can write $\gamma$ as\\
$\forall (x,\tilde{x})\in\cX\times\cX$,
\eqna{
\gamma(x,\tilde x) &= \int_{\cZ_{1:N}} p_{\theta_1}(\tilde x|z_1) \, q_{\phi_1}(z_{1:N}|x) \,p_D(x) \,dz_{1:N}  \\
&= \int_{\cZ_1} p_{\theta_1}(\tilde x|z_1) \, q_{\phi_1}(z_1|x) \,p_D(x) \,dz_1 \label{eq:app_transportplan_one}
}

The constraint in Eq.~\eqref{eq:marginal_constraints} on the $p_D$ marginal is trivially true as the integral over the second variable can be brought inside the integral over $\cZ_1$, after which all the integrals simply integrate to unity leaving $p_D$.

The constraint on the second marginal is obtained by integrating Eq.~\eqref{eq:app_transportplan_one} over the first variable,\\
$\forall{x}\in\cX$,
\eqna{
\int_{\cX} \gamma(x,\tilde x) \,dx 
&= 
\int_{\cX} \int_{\cZ_1} p_{\theta_1}(\tilde x|z_1)q_{\phi_1}(z_1|x) p_D(x)dz_1 \,dx \\
&= 
\int_{\cZ_1} p_{\theta_1}(\tilde x|z_1) \,\int_{\cX} q_{\phi_1}(z_1|x) p_D(x)dx \,dz_1 \\
\label{eq:app_second_marginal}
}
Thus, to satisfy Eq.~\eqref{eq:marginal_constraints}, we need:\\
$\forall{\tilde{x}}\in\cX$,
\eqn{
& \int_{\cZ_1} p_\Theta(\tilde x|z_1) \, \overbrace{\int_{\cX} q_{\phi_1}(z_1|x)p_D(x)dx}^{\equaldef \, q_1^\text{agg}(z_1)} \, dz_1 \overset{\text{need}}{=} \notag \\
& \hspace{3.5cm} \int_{\cZ_1} p_{\theta_1}(\tilde x|z_1) \, p_{\Theta_{2:N}}(z_1) \, dz_1
\label{eq:app_second_marginal_const}
}
where we use the generative model defined in Eq.~\eqref{eq:gen_model} and we introduced:\\
$\forall{z_1}\in\cZ_1$,
\eqn{
p_{\Theta_{2:N}}(z_1) \,\equaldef\,\int_{\cZ_{2:N}} \prod_{n=2}^N \, p_{\theta_n}(z_{n-1}|z_n) \,p(z_N) \, dz_{2:N}
}
To satisfy Eq.~\eqref{eq:app_second_marginal_const}, one obvious sufficient condition on the aggregated posterior distribution $Q_1^\text{agg}(Z_1)$ defined in Eq.~\eqref{eq:Qagg1} is Eq.~\eqref{eq:marginal_const}, namely that
\eqn{
\forall{z_1}\in\cZ_1, \quad q_1^\text{agg}(z_1) \,=\, p_{\Theta_{2:N}}(z_1)
\label{eq:app_marginal_const}  
}
which is what we sought out to show. However, Eq.~\eqref{eq:app_marginal_const} is a sufficient condition, not a necessary one: indeed Eq.~\eqref{eq:app_marginal_const} must only hold under $\int_{\cZ_1} P_\Theta(\tilde X|z_1)dz_1$. So for example, if $P_{\theta_1}(\tilde X|z_1)=P_{\theta_1}(\tilde X)$, then Eq.~\eqref{eq:app_second_marginal_const} would boil down to a constraint only on the expectations of $Q_1^\text{agg}(Z_1)$ and $P_{\Theta_{2:N}}(Z_1)$.

\subsection{WAE objective}
\label{app:wae_obj}

Starting from the definition of the OT distance given in Eq.~\eqref{eq:OT}, and using the WAE approach with density $\gamma$ written as Eq.~\eqref{eq:app_transportplan_one}, we find:
\eqn{
&\text{OT}_c(P_D,P_\Theta) 
\,=\, 
\underset{\Gamma\in\cP(P_D,P_\Theta)}{\inf} \, \int_{\cX\times\cX} \, c(x,\tilde x) \, d\Gamma(x,\tilde x)  \notag \\
\, &\leq \, \underset{
\substack{
Q_{\phi_1}(Z_1,Z_2,\ldots Z_n|X),
\\
\text{Eq.}\eqref{eq:app_second_marginal_const} \, \text{satisfied}
}
}{\inf} \, \int_{\cX\times\cX} c(x,\tilde x) \, \gamma(x,\tilde{x}) \, dx \, d\tilde{x} \notag
}
where $\gamma$ is defined in Eq.~\eqref{eq:app_transportplan_one}. Given that the above does not depend on $z_{>1}$, the inf can be written over $Q_{\phi_1}(Z_1|X)$ rather than the full $Q_{\phi_1}(Z_1,Z_2,\ldots Z_n|X)$. Replacing the constraint in the inf with the sufficient condition according to Eq.~\eqref{eq:marginal_const}, which amounts to replacing Eq.~\eqref{eq:app_second_marginal_const} with Eq.~\eqref{eq:app_marginal_const}, we obtain:
\eqn{
\text{OT}_c(P_D,P_\Theta) 
\,&\leq\,
\underset{
\substack{Q_{\phi_1}(Z_1|X), \\
Q_1^\text{agg}=P_{\Theta_{2:N}}
}
}{\inf} \; \int_{\cX\times\cX} c(x,\tilde x) \, \gamma(x,\tilde{x}) \, dx \, d\tilde{x}
\label{eq:app_OT}
}
with $\gamma$ still as in Eq.~\eqref{eq:app_transportplan_one}.
Eq.~\eqref{eq:wae_0} is then obtained by relaxing constraint in Eq.~\eqref{eq:app_OT}; replacing the hard constraint by a soft constraint via a penalty term added to the objective, weighted by a $\lambda_1$:
\eqn{
\widehat{W}_c(P_D,P_\Theta) \,&= \underset{Q_{\phi_1}(z_1|x)}{\inf} \Bigg[ \int_{\cX\times\cX} c(x,\tilde x) \, \gamma(x,\tilde{x}) \, dx \, d\tilde{x} 
\notag \\
& \hspace{1.25cm}
+\lambda_1 \, \cD_1\bigg( \, Q_1^\text{agg}(Z_1), \, P_{\Theta_{2:N}}(Z_1) \, \bigg) \Bigg] \label{eq:app_wae}
}
where $\cD_1$ is any divergence function between distributions on $\cZ_1$. 

\section{Experiments}
\label{app:exp}

\subsection{MNIST experiments}
\label{app:mnist}

\subsubsection*{Experimental setup}
\label{app:mnist_setup}
We train a deep-hierarchical latent-variable model with $N=5$ latent layers whose dimensions are $d_{\cZ_1}=32$, $d_{\cZ_2}=16$, $d_{\cZ_3}=8$, $d_{\cZ_4}=4$ and $d_{\cZ_5}=2$, respectively. We parametrise the generative and inference models as:
\eqn{
q_{\phi_i}(z_i|z_{i-1}) &= \mathcal{N}\big( z_i;\mu^q_i(z_{i-1}), \Sigma^q_i(z_{i-1}) \big), 
\quad i=1,\ldots, 5
\notag \\
p_{\theta_i}(z_{i-1}|z_i) &= \mathcal{N}\big( z_{i-1};\mu^p_i(z_i), \Sigma^p_i(z_i) \big),
\quad i=2,\ldots, 5
\notag \\
p_{\theta_1}(x|z_1) &= \delta\big( x - f_{\theta_1}(z_{1}) \big)
\label{eq_app:gaussian_params}
}
For both the encoder and decoder, the mean and diagonal covariance functions $\mu_i, \Sigma_i$ are fully-connected networks with 2 same-size hidden layers (consider $f_{\theta_1}$ as $\mu_1^p$). For $i=1,2,3,4,5$, the number of units is $2048 ,1024, 512, 256, 128$ respectively.

For the regularisation hyperparameters, we use $\prod_{i=1}^n\lambda_i = \lambda_\text{rec}^n / d_{z_n}$ for $n=1,\ldots,4$ (for each reconstruction term in the objective), and $\prod_{i=1}^5\lambda_i = \lambda_\text{match}$ (for the final divergence term). We then perform a grid search over the 25 pairs $(\lambda_\text{rec},\lambda_\text{match})\in\{0.005,0.01,0.05,0.1,0.5\}\otimes\{10^{-5},10^{-4},10^{-3},10^{-2},10^{-1}\}$ and find the best result (smallest Eq.~\eqref{eq:wae_unrolled_final}) is obtained with $(\lambda_\text{rec},\lambda_\text{match})=(0.01,10^{-4})$.

Finally, we choose the squared Euclidean distance as the cost function: $c_n(z_n,\tilde{z}_n)=\norm{z_n-\tilde{z}_n}^2$. The expectations in Eq.~\eqref{eq:wae_unrolled_final} are computed analytically whenever possible, and with Monte Carlo sampling otherwise.
We use batch normalisation \citep{BatchNorm} after each hidden fully-connected layer, followed by a ReLU activation \citep{relu}. We train the models over $5,000$ epochs using Adam optimiser \citep{ADAM} with default parameters and batch size of $128$.

\subsubsection*{Additional results}
\label{app:mnist_add_res}

\begin{figure}
\centering
\begin{subfigure}{.235\textwidth}
    \centering\includegraphics[width=\textwidth]{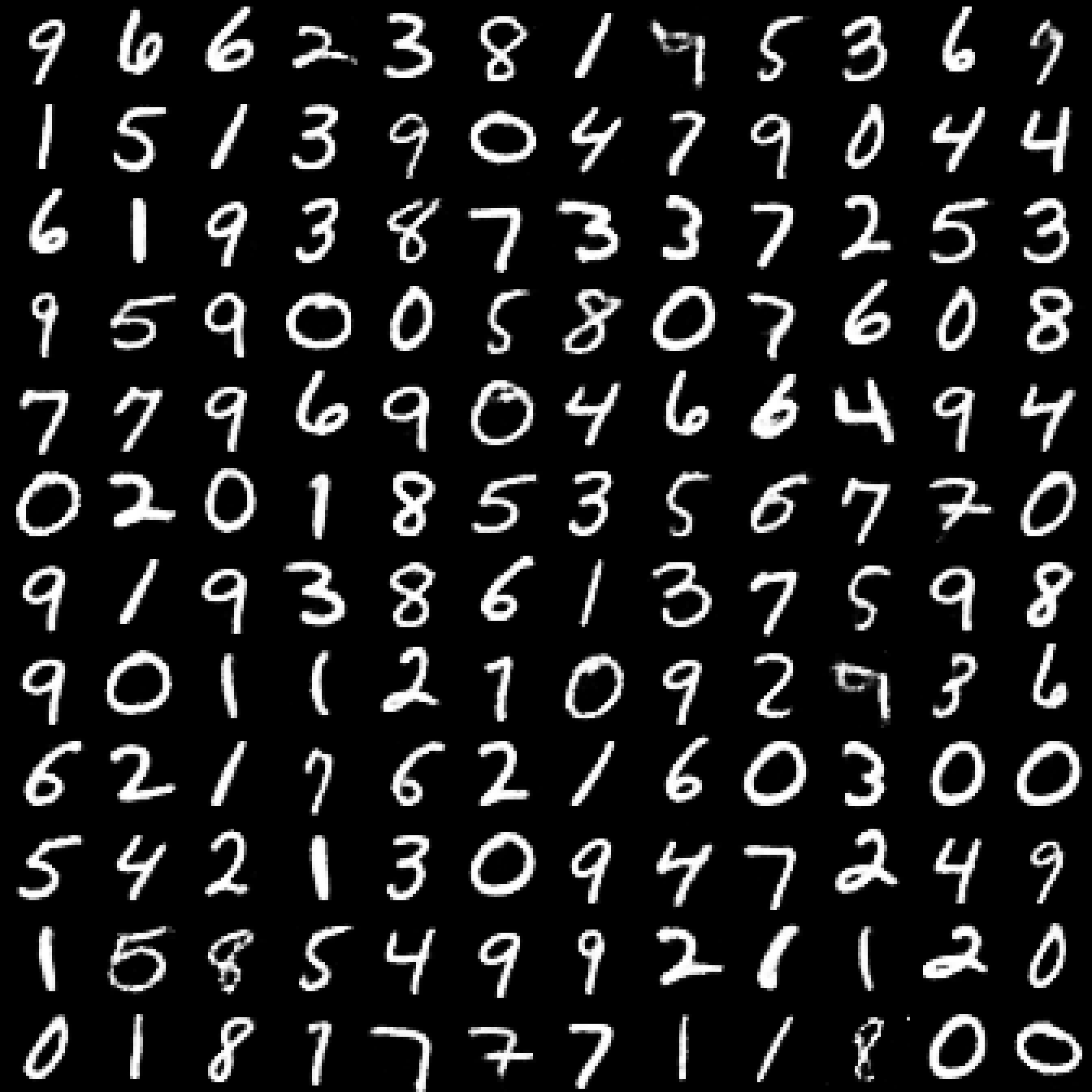}
    \vskip .13em
    \label{fig_mnist:app_prior_samples}
\end{subfigure}
\begin{subfigure}{.235\textwidth}
    \centering\includegraphics[width=\textwidth]{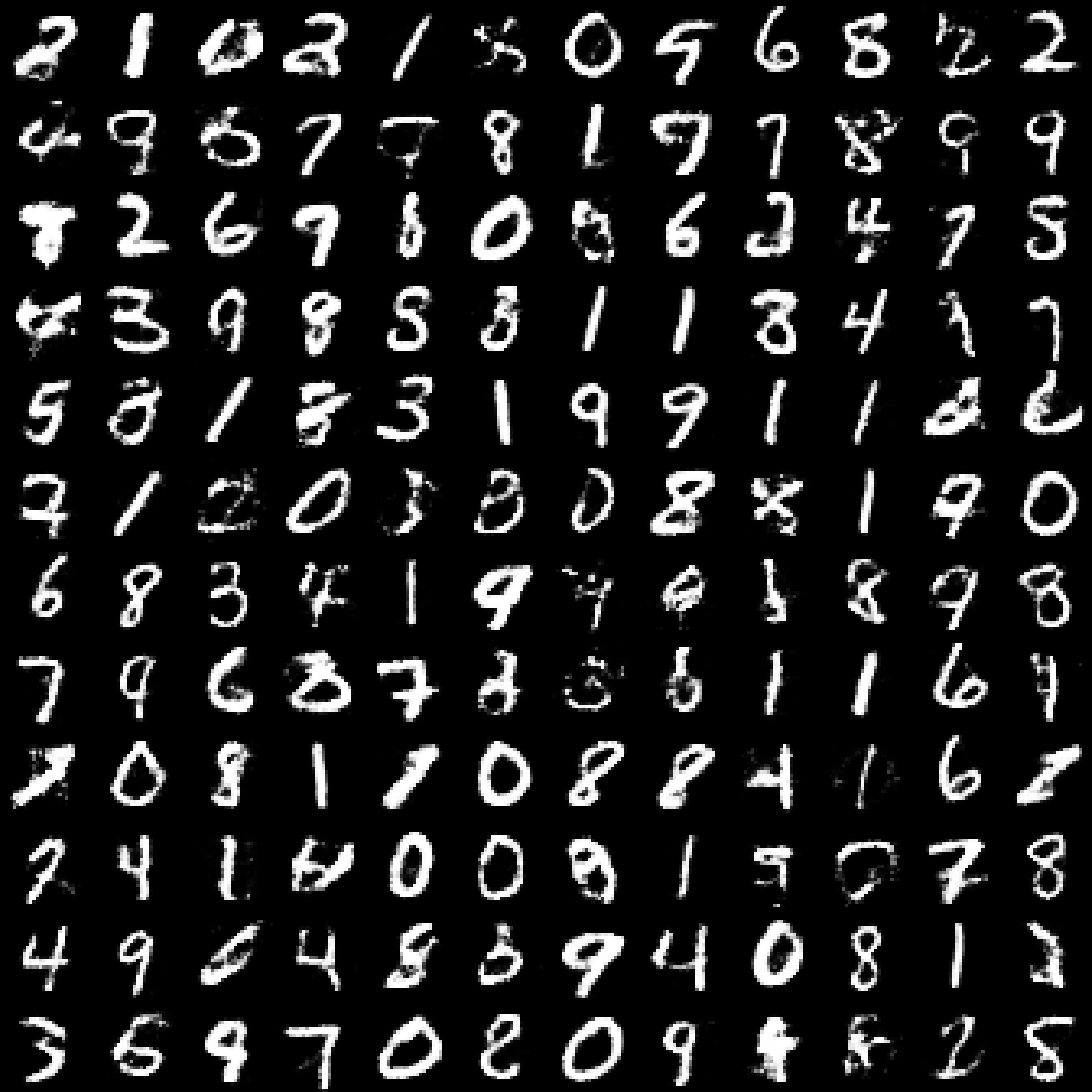}
    \vskip .13em
    \label{fig_mnist:app_imp_prior_samples}
\end{subfigure}
\begin{subfigure}{.235\textwidth}
    \centering\includegraphics[width=\textwidth]{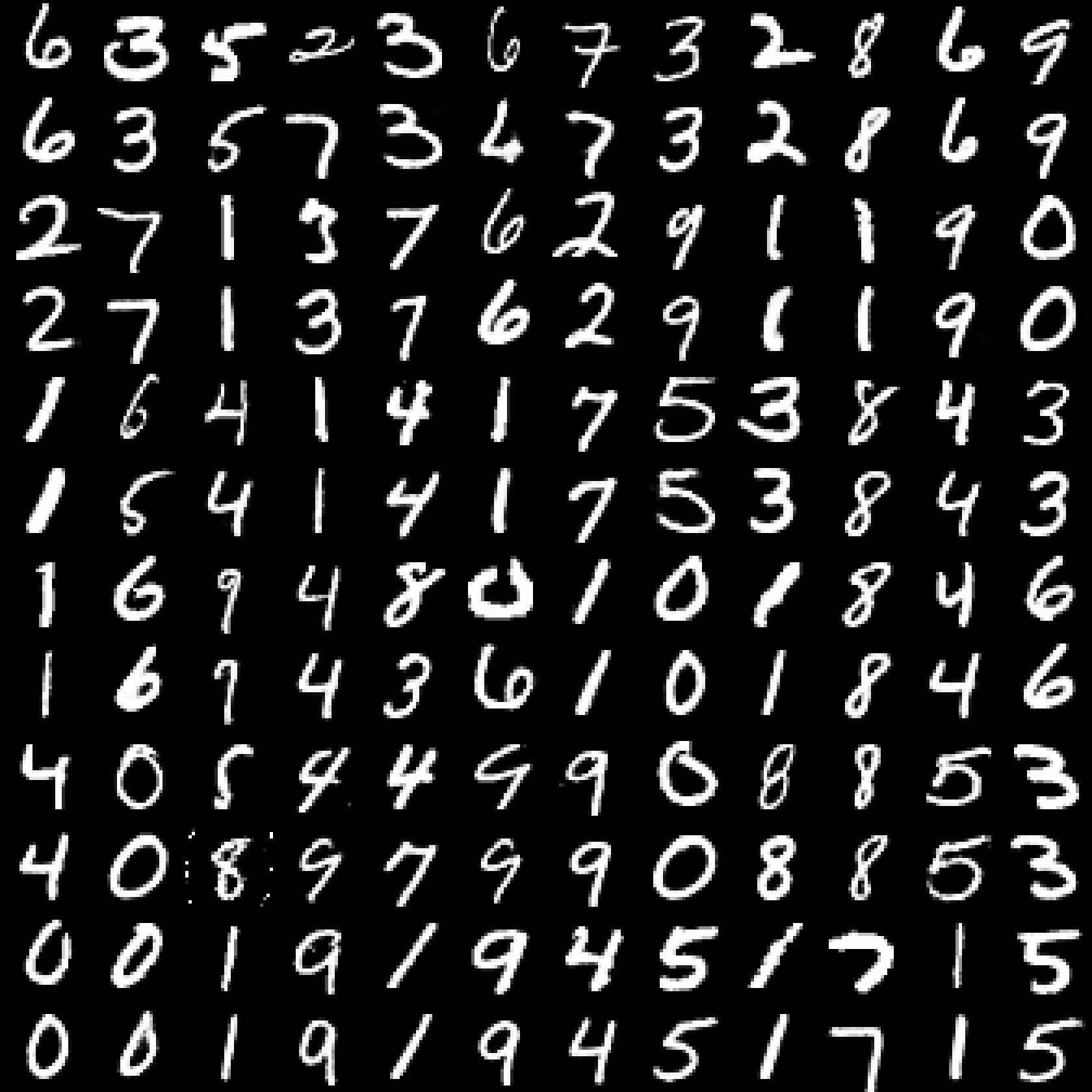}
    \vskip -.5em
    \label{fig_mnist:app_reconstructed}
\end{subfigure}
\begin{subfigure}{.235\textwidth}
    \centering\includegraphics[width=\textwidth]{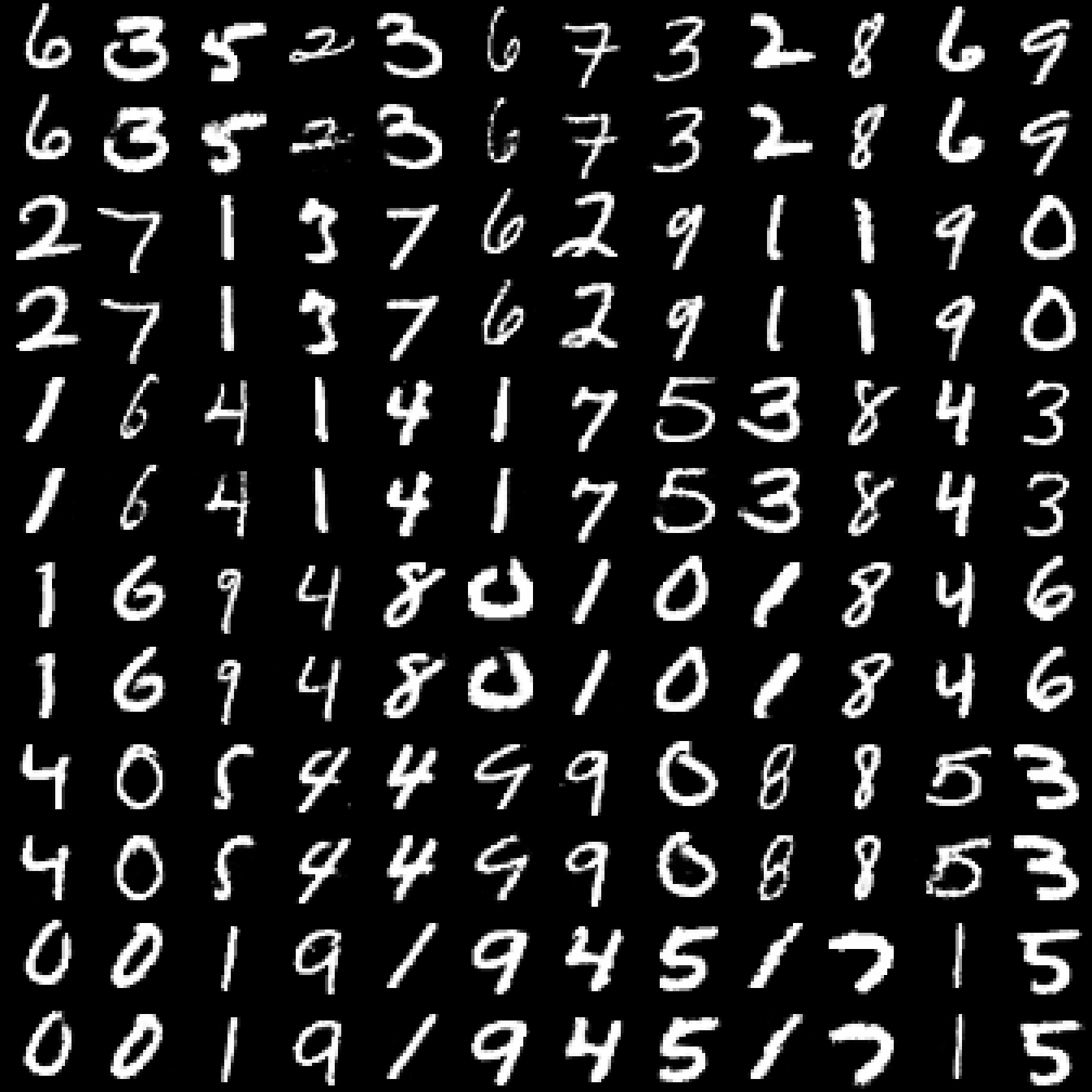}
    \vskip -.5em
    \label{fig_mnist:app_imp_reconstruced}
\end{subfigure}
\caption{5-layer \NAME{} (left column) \textit{versus} 1-layer implicit-prior WAE (right column). First row: Model samples.
Second row: Reconstructions (within pairs of rows, data is above with the corresponding reconstructions below).}
\label{fig_mnist:app_mnist}
\end{figure}

\begin{figure}
\begin{center}
\begin{subfigure}{.49\textwidth}
    \centering\includegraphics[width=\textwidth]{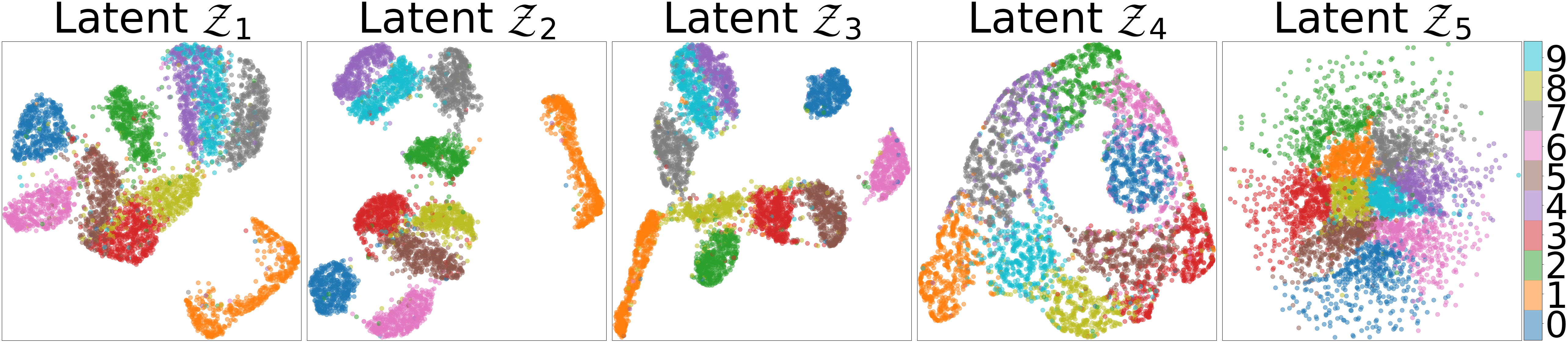}
\caption{}
\label{fig_mnist:umap}
\end{subfigure}\\
\vspace{1mm}
\begin{subfigure}{.49\textwidth}
    \centering\includegraphics[width=\textwidth]{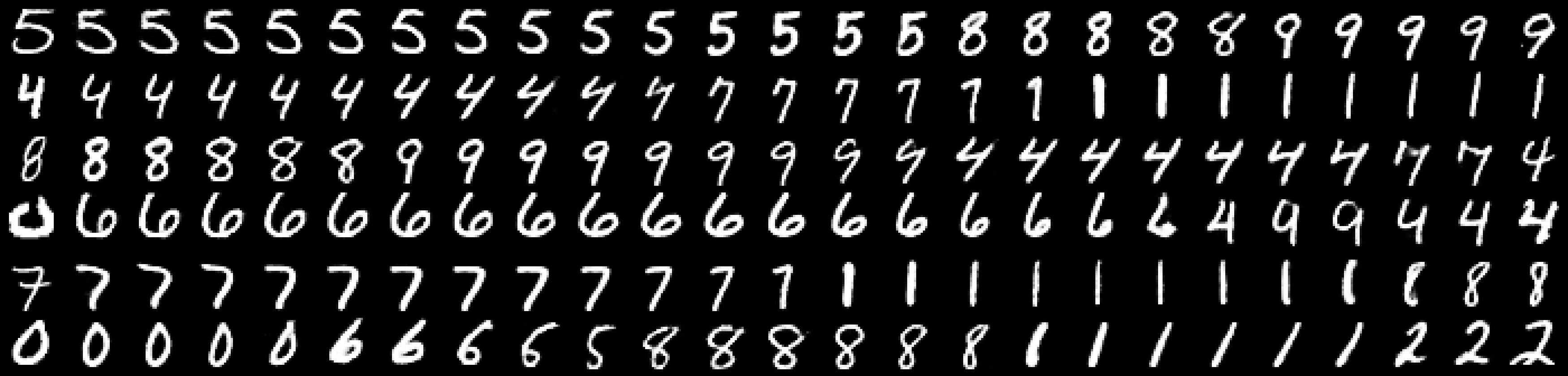}
\caption{}
\vskip -.5em
\label{fig_mnist:point_inter}
\end{subfigure}
\caption{5-layer \NAME. (a) UMAP \citep{2018arXivUMAP} visualisations of latent spaces $\mathcal{Z}_i$. Each colour corresponds to a digit label. $d_{\cZ_5}=2$ can be directly plotted; for higher dimensions we use UMAP. (b) Points interpolations; the first and last columns are actual data points with corresponding reconstructions shown in the second, respectively second-to-last, columns.}
\end{center}
\end{figure}

\subsection{Real word dataset}
\label{app:svhn_celeba}

\begin{figure}
\begin{center}
\begin{subfigure}{.4\textwidth}
    \centering
    \includegraphics[width=\textwidth]{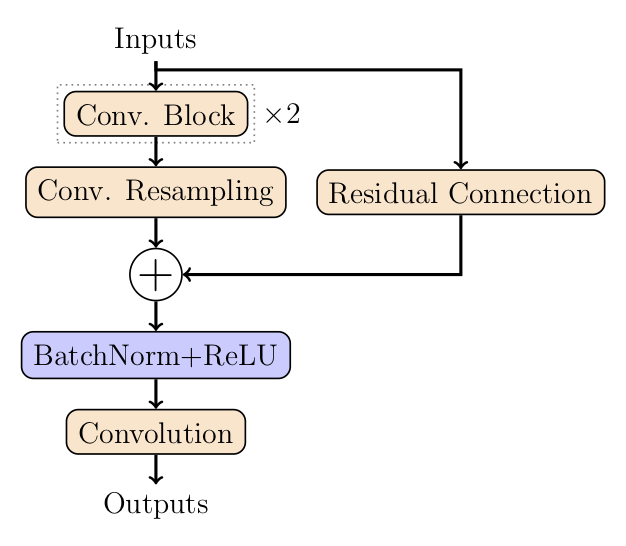}
    \setlength{\abovecaptionskip}{-15pt}
\end{subfigure}
\caption{Residual network with $3$ hidden convolutions.}
\label{fig:app_resnet}
\end{center}
\end{figure}

In the following experiments, the inference networks and generative models at each latent layer are taken to be Gaussian distributions with mean and covariance functions parametrised by 3-layer ResNet (similar to those of Eq.~\ref{eq_app:gaussian_params}).
A $M$-layer residual network is composed of $M-1$ convolutional blocks followed by a resampling convolution, and a residual connection. The outputs of the two are added and a last operation (either fully connected or convolution layer) is applied on the result. A convolutional block is composed of a convolution layer followed by batch normalisation \citep{BatchNorm} and a ReLU non-linearity \citep{relu}. We also use batch normalisation and ReLU after the sum of the convolutional blocks output and the residual connection. See  Figure~\ref{fig:app_resnet} for an example of a 3-layers residual network with a last convolution operation.
When resampling, we use a convolution layer with stride 2 in both the skip connection and the resampling covolution for the inference networks and a deconvolution layer with stride 2 in both the skip connection and the resampling convolution in the generative models. If no resampling is performed, then the resampling convolution is a simple convolution layer with stride 1 and the skip connection performs the identity operation. The latent dimensions are then given by the dimensions and the number of features in the last convolutional operation.

The ground cost is chosen to be the $L_2$-squared norm given by $c_n(z_n,\tilde{z}_n)=\norm{z_n-\tilde{z}_n}^2$, and the expectations in Eq.~\eqref{eq:wae_unrolled_final} are computed analytically whenever possible, otherwise using Monte Carlo sampling.

\subsubsection*{SVHN}
\label{app:svhn}

We train a 6-layers \NAME{} on the SVHN dataset using both the training dataset (73,257 digits) and the additional training dataset (531,131 digits). 
We train the models over $1000$ epochs using Adam optimiser \citep{ADAM} with default parameters and batch size of $100$.

As previously mentioned, we use 3-layer ResNet for the mean and covariance functions of the inference networks and generative models. More specifically, for each ResNet, we use $M=2$ convolutional blocks with the dimensions of the filters specified in the top-table of Figure~\ref{table:archi}.
Networks in layers $i=1, 2$ have $64$ convolution filters, layers $i=3,4$ $64$ and layers $5,6$ $128$, each filters having the same size within each residual network, and doubling (in the inference networks) or divide by 2 (in the generative models) their number in the resampling convolution if any resampling is performed. The latent layers $i=1, 3, 5$ have a stride of $2$ and we choose the number of features to be $2,1,2,1,2$ and $1$ for the latent layers $i=1\ldots6$ (top-table of Figure~\ref{table:archi} for the full details).

\begin{figure}
\begin{center}
\begin{subfigure}{.42\textwidth}
    \small
    \centering
    \renewcommand{\arraystretch}{1.4}
    \caption*{SVHN}
    \vskip -.4em
      \begin{tabular}{llll}
        \toprule[1.5pt]
        Layer i     & Filters dim.              & Resampling    & Output dim. \\
        \midrule       
        Layer 1     & 5$\times$5$\times$64      & down / up     & 16$\times$16$\times$2 \\
        Layer 2     & 3$\times$3$\times$64      & None          & 16$\times$16$\times$1 \\
        Layer 3     & 3$\times$3$\times$96      & down / up     & 8$\times$8$\times$2 \\
        Layer 4     & 3$\times$3$\times$96      & None          & 8$\times$8$\times$1 \\
        Layer 5     & 3$\times$3$\times$128     & down / up     & 4$\times$4$\times$2 \\
        Layer 6     & 3$\times$3$\times$128     & None          & 4$\times$4$\times$1 \\
        \bottomrule[1.5pt]
      \end{tabular}
    \setlength{\abovecaptionskip}{23pt}
    \label{table:svhn_archi}
\end{subfigure}\\
\vskip .8em
\begin{subfigure}{.42\textwidth}
    \small
    \centering
    \renewcommand{\arraystretch}{1.4}
    \caption*{CelebA}
    \vskip -.4em
      \begin{tabular}{llll}
        \toprule[1.5pt]
        Layer i     & Filters dim.              & Resampling    & Output dim. \\
        \midrule       
        Layer 1     & 7$\times$7$\times$64      & down / up     & 32$\times$32$\times$8 \\
        Layer 2     & 5$\times$5$\times$64      & None          & 32$\times$32$\times$6 \\
        Layer 3     & 3$\times$3$\times$64      & None          & 32$\times$32$\times$4 \\
        Layer 4     & 3$\times$3$\times$64      & down / up     & 16$\times$16$\times$8 \\
        Layer 5     & 3$\times$3$\times$96      & None          & 16$\times$16$\times$6 \\
        Layer 6     & 3$\times$3$\times$96      & None          & 16$\times$16$\times$4 \\
        Layer 7     & 3$\times$3$\times$96      & down / up     & 8$\times$8$\times$8 \\
        Layer 8     & 3$\times$3$\times$128     & None          & 8$\times$8$\times$6 \\
        Layer 9     & 3$\times$3$\times$128     & None          & 8$\times$8$\times$4 \\
        Layer 10    & 3$\times$3$\times$128     & down / up     & 4$\times$4$\times$8 \\
        \bottomrule[1.5pt]
      \end{tabular}
    \setlength{\abovecaptionskip}{23pt}
    \label{table:celeba_archi}
\end{subfigure}
\caption{Details of the architectures used in Section~\ref{sec:svhn_celeba}. Top-table: architecture of the 6-layer \NAME{} trained on SVHN. Bottom-table: architecture of the 10-layer \NAME{} trained on CelebA.}
\label{table:archi}
\end{center}
\end{figure}

\subsubsection*{CelebA}
\label{app:celeba}
We train a 10-layers \NAME{} on the CelebA dataset over $100$ epochs using Adam optimiser \citep{ADAM} with default parameters and batch size of $64$.

We use the same ResNet building blocks than previously, mainly, $M=2$ convolutional block, each filter within a ResNet block having the same size, and doubling their number in the last convolutional operation. Networks in layers $i=1, 2, 3, 4$ have $64$ filters, $96$ in layers $i=5,6,7$ and $128$ in the remaining layers. The latent layers $i=1, 4, 7, 10$ have stride $2$ and we choose the number of features to be $8,4,2,8,4,2,8$ (See bottom-table of Figure~\ref{table:archi} for the full details).

\end{document}